\documentclass[letterpaper]{article} 
\usepackage{aaai24}  
\usepackage{times}  
\usepackage{helvet}  
\usepackage{courier}  
\usepackage[hyphens]{url}  
\usepackage{graphicx} 
\usepackage{natbib}  
\usepackage{caption} 
\usepackage{algorithm}
\usepackage{algorithmic}

\usepackage{subcaption}
\usepackage{multirow}
\usepackage{pifont}
\usepackage{arydshln}
\usepackage{color}
\usepackage{booktabs}
\usepackage{threeparttable}
\usepackage{makecell}
\usepackage{amsmath}
\usepackage{amssymb}
\def\modelname{Whiten-MTD}
\newcommand{\eg}{\emph{e.g.,}~}
\newcommand{\ie}{\emph{i.e.,}~}
\newcommand{\etal}{\emph{et al.}~}

\newcommand{\etc}{\emph{etc.}~}

\usepackage{newfloat}
\usepackage{listings}
\DeclareCaptionStyle{ruled}{labelfont=normalfont,labelsep=colon,strut=off} 
\floatstyle{ruled}
\newfloat{listing}{tb}{lst}{}
\floatname{listing}{Listing}
\title{Let All be Whitened: Multi-Teacher Distillation for Efficient Visual Retrieval}
\author{
    Zhe Ma\textsuperscript{\rm 1},
    Jianfeng Dong\textsuperscript{\rm 2,\rm 4}\footnote{Corresponding authors.},
    Shouling Ji\textsuperscript{\rm 1},
    Zhenguang Liu\textsuperscript{\rm 1}\footnotemark[1],
    Xuhong Zhang\textsuperscript{\rm 1},
    Zonghui Wang\textsuperscript{\rm 1}\footnotemark[1],\\
    Sifeng He\textsuperscript{\rm 3},
    Feng Qian\textsuperscript{\rm 3},
    Xiaobo Zhang\textsuperscript{\rm 3},
    Lei Yang\textsuperscript{\rm 3}
}
\affiliations {
    \textsuperscript{\rm 1}Zhejiang University,
    \textsuperscript{\rm 2}Zhejiang Gongshang University,
    \textsuperscript{\rm 3}Ant Group,
    \textsuperscript{\rm 4}Zhejiang Key Lab of E-Commerce\\
    \{mz.rs, sji, zhangxuhong, zhwang\}@zju.edu.cn,
    \{dongjf24, liuzhenguang2008, hsf215kg\}@gmail.com,\\
    \{youzhi.qf, yl149505\}@antgroup.com,
    ayou.zxb@antfin.com
}

\usepackage{bibentry}

\begin{document}

\maketitle

\begin{abstract}
Visual retrieval aims to search for the most relevant visual items, e.g., images and videos, from a candidate gallery with a given query item. Accuracy and efficiency are two competing objectives in retrieval tasks. Instead of crafting a new method pursuing further improvement on accuracy, in this paper we propose a multi-teacher distillation framework Whiten-MTD, which is able to transfer knowledge from off-the-shelf pre-trained retrieval models to a lightweight student model for efficient visual retrieval. Furthermore, we discover that the similarities obtained by different retrieval models are diversified and incommensurable, which makes it challenging to jointly distill knowledge from multiple models. Therefore, we propose to whiten the output of teacher models before fusion, which enables effective multi-teacher distillation for retrieval models. Whiten-MTD is conceptually simple and practically effective. Extensive experiments on two landmark image retrieval datasets and one video retrieval dataset demonstrate the effectiveness of our proposed method, and its good balance of retrieval performance and efficiency. Our source code is released at https://github.com/Maryeon/whiten\_mtd.
\end{abstract}

\section{Introduction}
\label{sec:intro}
Visual retrieval, such as image retrieval and video retrieval, is a long-standing problem in the computer vision community, which aims to search for the most similar items to a given query from a large number of candidates~\cite{radenovic2018fine,revaud2019learning,noh2017large,yang2021dolg}.
It supports a wide range of applications including instance matching~\cite{cao2020unifying,radenovic2018fine,revaud2019learning}, fine-grained recognition~\cite{dong2021fine}, product recommendation~\cite{kim2021dual,dong2018feature}, etc.

Recent visual retrieval methods have been dominated by the Deep Neural Networks~(DNNs)~\cite{he2016deep,vaswani2017attention} based solutions.
Given an overview of recent developments on visual retrieval problems~\cite{chen2022deep,shen2020advance}, there are two considerations that must be taken into account.
On the one hand, with the large amounts of pre-trained retrieval models being publicly available, it motivates us to make a complementary integration of them, in other words, take their essence and discard the dregs.
On the other hand, despite the significant performance improvement boosted by DNNs, a major concern is their heavy overhead of computation. A compact and efficient retrieval model is required for large-scale scenarios.
In this work, we jointly take the two considerations by aggregating multiple cumbersome pre-trained retrieval models into a more efficient one. 
To this end, we resort to knowledge distillation \cite{hinton2015distilling,passalis2018learning,beyer2022knowledge}, which is a technique proposed to transfer the knowledge from a large model or an ensemble of large models (teachers) to a small model (student) to reduce computation overhead, without significant loss in performance. 
We propose a multi-teacher distillation framework for efficient visual retrieval, where off-the-shelf pre-trained retrieval models are regarded as teachers and distilled to a more lightweight student model.

\begin{figure}[t]
  \centering
  \begin{subfigure}{0.45\linewidth}
    \includegraphics[width=\linewidth]{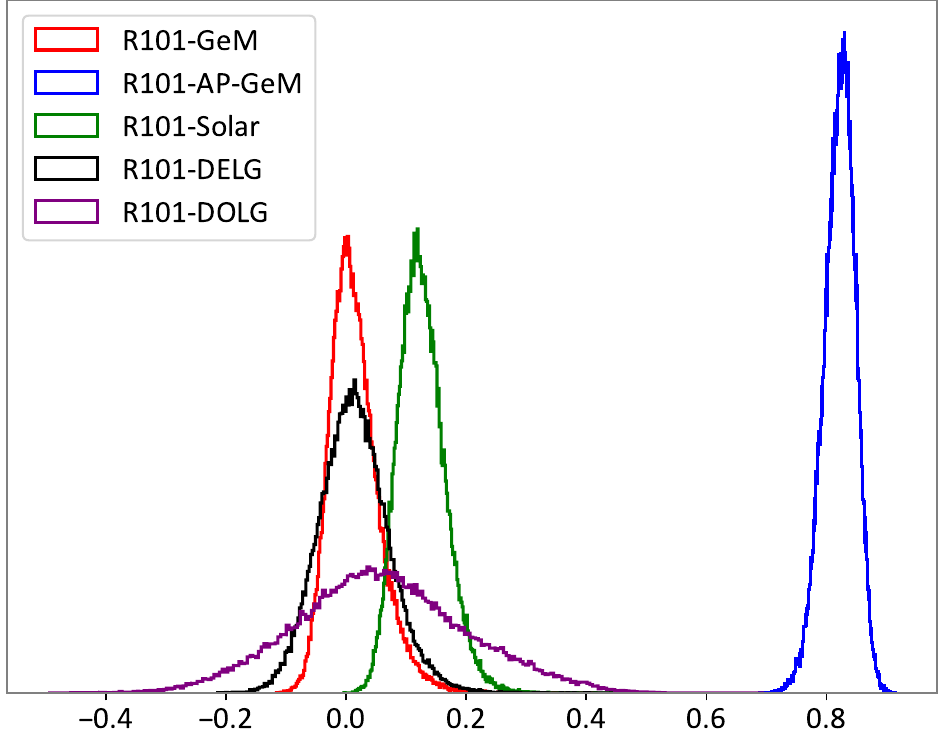}
    \caption{Without whitening.}
    \label{fig:sim_dist_a}
  \end{subfigure}
  \hfill
  \begin{subfigure}{0.45\linewidth}
    \includegraphics[width=\linewidth]{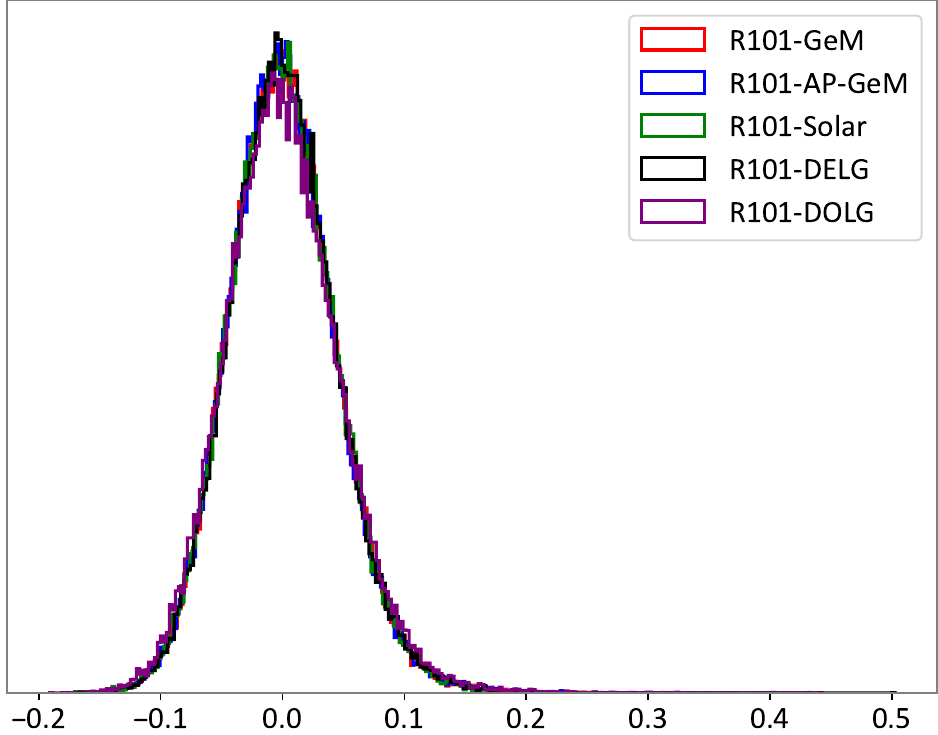}
    \caption{With whitening.}
    \label{fig:sim_dist_b}
  \end{subfigure}
  \caption{Cosine similarity distributions of five existing landmark image retrieval models on GLDv2-clean dataset. Their similarity distributions are clearly different, making the similarities between different models incommensurable.
  Such discrepancy can be alleviated through whitening their output features.
  }
  \label{fig:sim_dist}
\end{figure}

The keys to multi-teacher distillation are the \textit{knowledge} to be transferred from teachers to a student, and the \textit{fusion strategy} to combine the knowledge from teachers. 
The knowledge used in existing works can be categorized into feature-based~\cite{hinton2015distilling,romero2014fitnets,beyer2022knowledge} and relation-based~\cite{yim2017gift,tian2019contrastive,park2019relational,passalis2020heterogeneous}. Feature-based knowledge distillation constrains the student's intermediate or final outputs to be aligned to those of teacher models, forming a point-to-point fitting. Relation-based knowledge distillation typically transfers the similarity between layers \cite{yim2017gift,passalis2020heterogeneous} or samples \cite{park2019relational,tian2019contrastive} to the student model. As retrieval is generally modeled as a ranking problem where similarity scores defined between samples are used to sort the candidates,
we employ the similarity between samples as the knowledge to be transferred.

Another key problem of multi-teacher distillation is how to fuse the knowledge of teacher models. 
Previous multi-teacher distillation works are mainly developed for classification models \cite{bucilua2006model,hinton2015distilling,shen2019meal,yuan2021reinforced}, whose knowledge is conveyed by the output categorical probabilistic distributions that has equivalent measure and are commensurable between different models. Hence, simple strategies like average \cite{bucilua2006model,hinton2015distilling}, random selection \cite{shen2019meal} or weighted summation \cite{yuan2021reinforced} can be reasonably utilized.
However, due to the nature of retrieval methods that impose relational constraints on samples rather than push them to a finite number of prototypes (weights of the last classification layer) as in normal classification models, the representation space obtained by different visual retrieval models will be diversified and not commensurable.
Figure \ref{fig:sim_dist_a} shows the cosine similarity distribution of five existing landmark image retrieval models, demonstrating the clear discrepancy between retrieval models. 
Therefore, simply and directly utilizing the fusion strategy defined for image classification is suboptimal for visual retrieval.
How to fuse the knowledge of multiple teachers becomes a unique problem for retrieval tasks.

To alleviate the discrepancy of retrieval models, we propose to firstly align the similarity distributions of different retrieval models before any fusion operations.
To this end, we propose to utilize whitening, which transforms teacher models' output into a spherical distribution. We will show that under spherical distribution, similarity distributions of teacher models can be perfectly aligned and more commensurable (as exemplified in Figure \ref{fig:sim_dist_b}), without sacrificing their distillation performance.
Additionally, we devise multiple heuristic fusion strategies to aggregate predictions of teacher models. The fusion strategies compare and integrate the pairwise similarities calculated by each teacher model and produce the final knowledge to be transferred to the student model.

Our key contributions can be summarized as follows:
\begin{itemize}
\item We propose a simple but effective \textbf{Whiten}ing-based \textbf{M}ulti-\textbf{T}eacher \textbf{D}istillation (\modelname) framework for visual retrieval. \modelname~is able to readily transfer knowledge from multiple strong and cumbersome retrieval models into a lightweight one.
To the best of our knowledge, this is the first multi-teacher distillation work for visual retrieval.

\item In order to make different retrieval models commensurable, we propose to whiten the output of teacher models to align the discrepant similarity distribution.
Besides, we devise five heuristic fusion strategies, and demonstrate which one is best suited for our multi-teacher distillation framework by empirical study.

\item We conduct extensive experiments on two kinds of visual retrieval tasks, \ie instance image retrieval and video retrieval. Compared to the state-of-the-art \cite{song2023boosting,lee2022correlation,kordopatis2022dns} that are based on ResNet-50, ResNet-101 \cite{he2016deep} or ViT-B \cite{dosovitskiy2021an} and using the reranking strategy, we are able to achieve comparable performance without the reranking using more lightweight ResNet-18 or ResNet-34.

\end{itemize}

\section{Related work}
\label{sec:rel_work}

\subsection{Visual retrieval}
We review two specific fields of visual retrieval, \ie instance image retrieval, and near-duplicate video retrieval.

The goal of instance image retrieval is to search for images containing the same instance (\eg landmark, person) as the query. Earlier efforts are based on hand-crafted features
while recent works \cite{dusmanu2019d2,noh2017large,luo2019contextdesc} tend to utilize DNN features. For DNN features, local features encode rich regional information and are usually used to perform geometric verification \cite{fischler1981random}. However, the local feature matching is time-consuming and always adopted in the reranking process. Global features summarize the images and are more compact. They are obtained by directly taking the activations after fully connected layer or pooling the feature maps after convolutional layer. Various pooling operators such as R-MAC \cite{tolias2015particular}, GeM \cite{radenovic2018fine}, are proposed to capture regional discriminative information.
In addition, there are other works focusing on fine-tuning pre-trained classification models for instance image retrieval, including designing loss functions \cite{revaud2019learning}, sampling strategies \cite{arandjelovic2016netvlad,gordo2016deep}, etc.

Near-duplicate video retrieval \cite{shen2020advance,liu2013near} focuses on video matching to construct robust video representations against severe transformations.
Recent methods \cite{kordopatis2017near1,kordopatis2017near2,kordopatis2019visil,shao2021temporal,he2021self} all resort to DNNs. 
Based on pre-trained models to extract frame features, they focus on designing upper frame feature aggregation schemes, such as Bag-of-Words \cite{kordopatis2017near1}, Deep Metric Learning \cite{kordopatis2017near2}, Transformer \cite{shao2021temporal}, \etc
Computation efficiency is much more crucial for videos, which usually contain tens of thousands of images.

Current state-of-the-art visual retrieval methods achieve their superior performance with deep neural networks, such as ResNet-101, pursuing the maximization of accuracy. We instead focus on the efficiency of the underlying backbones.

\subsection{Distillation for visual retrieval}
Knowledge distillation is a technique that was initially designed to improve model efficiency \cite{bucilua2006model,hinton2015distilling}.
Knowledge distillation has been widely explored in representation learning \cite{caron2021emerging,noroozi2018boosting}, object detection \cite{zhixing2021distilling,kang2021instance,guo2021distilling}, action recognition \cite{liu2023lite}, etc.

For visual retrieval which depends on similarity between samples, relation-based knowledge distillation methods can be felicitously exploited \cite{passalis2018learning,fang2021seed,dong2023dual}. What's more, a specific framework called asymmetric retrieval has also been proposed to mitigate the accuracy-efficient trade-off by processing database samples with a large model while processing online queries with a lightweight one. Approaches \cite{wu2022contextual,duggal2021compatibility} in this framework typically distill the large model into a lightweight one and deploy it online.

Existing work for visual retrieval belongs to a single-teacher distillation setting while multi-teacher distillation is mainly studied in the context of image classification \cite{bucilua2006model,ba2014deep,you2017learning,shen2019meal,yuan2021reinforced}. Fusion strategies of teacher models are the central topic of this problem.
Averaging of all teacher models' predictions is a normal choice \cite{bucilua2006model,ba2014deep}.
Other strategies include voting \cite{you2017learning}, random selection \cite{shen2019meal}, selection by reinforcement learning agent \cite{yuan2021reinforced}, \etc.

Overall, the problem of multi-teacher distillation for visual retrieval is not well addressed and the intractable discrepancy between visual retrieval models hinders the direct usage of existing multi-teacher distillation methods.

\section{Method}
\label{sec:method}
We expect to transfer the knowledge from multiple teacher models, typically strong but cumbersome, to a lightweight student model.
After the distillation, the teacher models are discarded, and only the student model is retained for efficient visual retrieval.
Without loss of generality, we formulate our method in the context of image retrieval, which can be easily adapted to other visual retrieval tasks.

In what follows, we first describe the basic similarity-based single-teacher distillation. Then we extend it to multi-teacher distillation based on aggregating similarity matrices output by different teacher models and introduce five heuristic similarity fusion strategies. Finally we propose to use whitening to eliminate the similarity discrepancy among teacher models, which enables more effective and stable multi-teacher distillation.

\begin{figure}[tbp]
   \centering
   \includegraphics[width=0.9\linewidth]{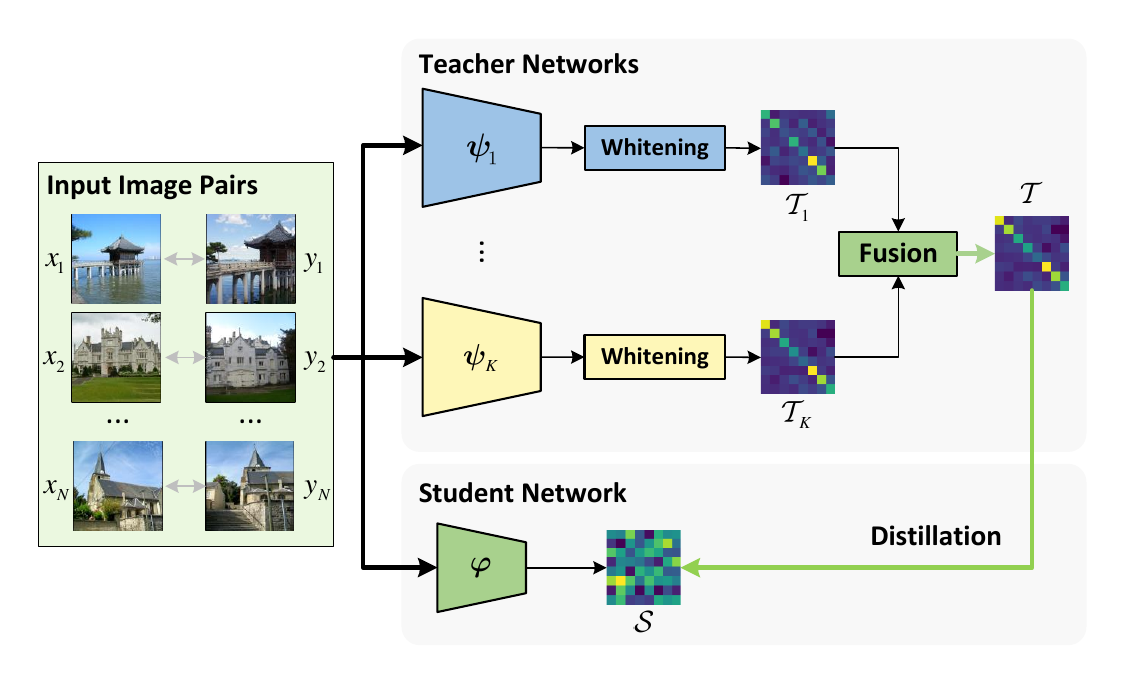}
   \caption{An overview of our multi-teacher distillation framework.
   Due to the similarity discrepancy of different retrieval models, the whitening is utilized on each teacher model to make their similarity commensurable. The fused similarity map as the knowledge of multiple teacher models is transferred to a lightweight student model.
   After distillation, the teacher models are discarded, and only the student model is retained for efficient visual retrieval.
   }
   \label{fig:method}
\end{figure}

\subsection{Single-teacher distillation}
Existing single-teacher distillation works \cite{passalis2018learning,fang2021seed,wu2022contextual} are analogous but only differ in the exact similarities used for distillation. For a clear validation of our multi-teacher distillation approach, we adopt the basic distillation method, although some advanced components can also be included, such as a queue structure \cite{fang2021seed}.

Specifically, both teacher and student models are image encoders that embed images from a high-dimensional input space  $\mathbb{R}^m$ into a low-dimensional representation space.
We denote the teacher model as $\psi(\cdot):\mathbb{R}^m\to\mathbb{R}^{n_t}$, and the student model as $\varphi(\cdot):\mathbb{R}^m\to\mathbb{R}^{n_s}$, where $n_t$ and $n_s$ indicate the output dimension of the teacher model and student model respectively. Suppose $\psi(\cdot)$ and $\varphi(\cdot)$ are both $l_2$-normalized functions, \ie $\Vert \psi(\cdot) \Vert_2=1$ and $\Vert \varphi(\cdot) \Vert_2=1$.
Let $\mathcal{B} = \{(x_i, y_i)\}^N_{i=1}$ be a batch of relevant positive image pairs.
Accordingly, $x_i$ and $y_j$ ($i\neq j$) are negative, meaning that they are semantically irrelevant.
The image pairs can be sampled either in a supervised or unsupervised way.
For example, in landmark instance image retrieval, according to the annotation, two images containing the same landmark are regarded as a positive pair, otherwise a negative pair. For video retrieval, they are obtained by image augmentation as widely used in self-supervised learning \cite{grill2020bootstrap,caron2021emerging,chen2021empirical}.

We define cosine similarity as a measure of semantic relation between images, then a similarity matrix $\mathcal{T}\in\mathbb{R}^{N\times N}$ can be calculated using the teacher model $\psi$, where $\mathcal{T}[i,j]=\psi(x_i)^\top \psi(y_j)$.
Similarly, $\mathcal{B}$ is represented by the student model as a similarity matrix $\mathcal{S}\in\mathbb{R}^{N\times N}$, where $\mathcal{S}[i,j]=\varphi(x_i)^\top \varphi(y_j)$.
Moreover, we convert the similarity matrices $\mathcal{T}$ and $\mathcal{S}$ into probabilistic similarity matrices $\mathcal{Q}$ and $\mathcal{P}$ which enhances the relativity between similarity scores rather than their absolute values. Specifically, each row of the similarity matrices is normalized into a probabilistic distribution using Gaussian kernel with a temperature:
\begin{equation}
  \mathcal{P}[i,j]=\frac{e^{\mathcal{S}[i,j]/\tau}}{\sum_{k=1}^Ne^{\mathcal{S}[i,k]/\tau}},
  \mathcal{Q}[i,j]=\frac{e^{\mathcal{T}[i,j]/\tau}}{\sum_{k=1}^Ne^{\mathcal{T}[i,k]/\tau}}
  \label{eq:kde}
\end{equation}
where $\tau$ is the temperature factor.
Finally the Kullback-Leibler (KL) divergence is minimized between each row of teacher probabilistic similarity matrix $\mathcal{Q}$ and student probabilistic similarity matrix $\mathcal{P}$ to conduct the distillation:
\begin{equation}
  \mathcal{L}=\frac{1}{N}\sum_i^N KL\left(\mathcal{P}_i||\mathcal{Q}_i\right).
  \label{eq:loss}
\end{equation}

\subsection{Multi-teacher distillation}
We now extend to multi-teacher distillation by considering $K>1$ teacher models $\psi_k(\cdot)\in\mathbb{R}^m\to\mathbb{R}^{n_{t_k}}$.
To combine the knowledge of multiple teacher models, we propose to fuse $K$ similarity matrices $\mathcal{T}_k$ produced by the teacher models into an aggregated similarity matrix $\mathcal{T}$ using a fusion strategy function $\mathcal{F}$,
\begin{equation}
  \mathcal{T}=\mathcal{F}(\mathcal{T}_1,\mathcal{T}_2,...,\mathcal{T}_K).
  \label{eq:fusion}
\end{equation}
Although elaborate strategies can be designed, we study five heuristic fusion strategies which are shown to be effective in our experiments. All the fusion strategies perform element-wise comparison and selection.
For each location $(i,j)$, we first include two normal strategies, which are:

$\bullet$ \textit{mean}: $\mathcal{T}[i,j]=\frac{1}{K}\sum_{k=1}^K\mathcal{T}_k[i,j]$.

$\bullet$ \textit{rand}: $\mathcal{T}[i,j]=\mathcal{T}_r[i,j],r\in [1,K]$.

Additionally, we design other three strategies based on decoupling the similarity matrix into inner-class similarities~(diagonal) and inter-class similarities~(off-diagonal), which are  \textit{max-min}, \textit{max-mean}, and \textit{max-rand}. All these strategies choose to take the maximum of similarities on the diagonal, favoring more compact inner-class clustering, while differing in the choice on the off-diagonal. For off-diagonal similarities (\ie $i\neq j$), the rules are:

$\bullet$ \textit{max-min}: $\mathcal{T}[i,j]=\min\limits_{k\in [1,K]}(\mathcal{T}_k[i,j])$.

$\bullet$ \textit{max-mean}: $\mathcal{T}[i,j]=\frac{1}{K}\sum_{k=1}^K\mathcal{T}_k[i,j]$. 

$\bullet$ \textit{max-rand}: $\mathcal{T}[i,j]=\mathcal{T}_r[i,j],r\in [1,K]$.

However, these fusion strategies are not sufficient for effective multi-teacher distillation unless coupled with teacher model whitening, which will be described next.

\subsection{Eliminating teacher discrepancy}
\label{sec:pca-whitening}
The fusion strategy $\mathcal{F}$ in Eq. \ref{eq:fusion} requires to make comparison between different teachers and then perform a selection. However, due to the ranking property of retrieval, only the relative order of samples is considered regardless of the absolute value of similarities. 
This enables so much flexibility for models' outputs that models with similar retrieval performance may have diverse output similarity distributions. 
This can be demonstrated from a statistical visualization of the similarity distributions of different teacher models, as depicted in Figure \ref{fig:sim_dist_a}. If the distribution gap is neglected, existing normal multi-teacher fusion strategies, such as averaging, will be less effective as the output will be biased to the extreme distribution of a specific teacher model.

To bridge the gap among different teacher models, we propose to whiten their output representations before calculating similarity scores. Whitening is a linear transformation that transforms the representations into spherical distribution. Followed by $l_2$-normalization, the resulting representations will distribute uniformly on the unit sphere surface \cite{ermolov2021whitening}. As proved by \citeauthor{cai2013distributions} \etal \shortcite{cai2013distributions}, the angle $\theta$ between two independent random vectors distributed uniformly on the unit sphere surface in $\mathbb{R}^n$ converges to a distribution with the probability density function $f(\theta) = \frac{1}{\sqrt{\pi}}\cdot\frac{\Gamma(\frac{n}{2})}{\Gamma(\frac{n-1}{2})}\cdot(\sin\theta)^{n-2},\theta\in\left[ 0,\pi\right]$.
Therefore, cosine similarities of different teacher models calculated with the whitened representations will follow the same distribution (see statistical results in Figure \ref{fig:sim_dist_b}). Whitening has been applied in recent works as a post-processing technique for further performance improvement in instance matching \cite{cao2020unifying,revaud2019learning} or normalization technique to prevent representation collapse in self-supervised learning \cite{ermolov2021whitening,weng2022investigation}. We otherwise uncover the capability of whitening to align teachers' similarity distributions, which is crucial for multi-teacher distillation.

Specifically, we adopt PCA-whitening which is learned on the training set and fixed in the distillation stage. Given a training set $\mathcal{D}=\{x_i\}_{i=1}^M$ where $M$ is the total number of samples, for the $k$-th teacher, the PCA-whitening $\mathcal{W}_k$ is a linear transformation $W_k$:
\begin{equation}
\mathcal{W}_k(\psi_k(\cdot))=W_k\left(\psi_k(\cdot)-b_k\right),
  \label{eq:whiten}
\end{equation}
where $W_k\in\mathbb{R}^{n_c\times n_{t_k}}$, $b_k\in\mathbb{R}^{n_{t_k}}$, $n_c$ is the dimensionality of whitened representations. $b_k=\frac{1}{M}\sum_{i=1}^M\psi_k(x_i)$ is the mean representation of all the training samples and $W_k$ is the eigen-system of $\Sigma$ satisfying $W_k^\top W_k=\Sigma^{-1}$, where $\Sigma$ is the covariance matrix of zero-meaned $\psi_k(\cdot)$ (\ie $\psi_k(\cdot)-b_k$). We provide a detailed discussion on the efficacy of whitening and the choice of $n_c$ in supplementary materials.

\section{Experiments}
\label{sec:exp}

To verify the viability of \modelname, we evaluate it on two common visual retrieval tasks, \ie landmark image retrieval and near-duplicate video retrieval. We provide implementation details and additional experiment results in supplementary materials.

\subsection{Landmark image retrieval}

\subsubsection{Datasets}
Following previous works \cite{lee2022correlation,song2023boosting}, we use the clean version of Google Landmark Dataset v2 (GLDv2-clean) \cite{weyand2020google} as the training set and two additional independent datasets RParis6k ($\mathcal{R}$Par) and ROxford5k ($\mathcal{R}$Oxf) \cite{radenovic2018revisiting} for evaluation.
GLDv2-clean consists of around 1.6 million images of 80K landmark instances.
$\mathcal{R}$Oxf and $\mathcal{R}$Par are the revisited version of Paris6k \cite{philbin2008lost} and Oxford5k \cite{philbin2007object} building datasets. The revisited versions improved the reliability of ground truth and introduced three new protocols of varying difficulty, \ie Easy, Medium and Hard.  Both $\mathcal{R}$Par and $\mathcal{R}$Oxf contains 70 queries, and 6,322 and 4,993 gallery images respectively.
Additionally, 1M distractor images are included as the extra gallery images for large-scale evaluation on both datasets.

\subsubsection{Metrics}
As done in \cite{radenovic2018fine,revaud2019learning}, we report mAP under the Medium and Hard evaluation protocols on $\mathcal{R}$Par and $\mathcal{R}$Oxf, as well as large-scale evaluation with the 1M distractors.
For the model complexity, we report the model size in terms of the number of parameters and computation overhead during the inference.
The computation overhead is measured by the number of GFLOPs when a model encodes a given image of size $1024\times 768$.

\subsubsection{Networks}
We adopt five pre-trained models, \ie R101-GeM \cite{radenovic2018fine}, R101-AP-GeM \cite{revaud2019learning}, R101-SOLAR \cite{ng2020solar}, R101-DELG \cite{cao2020unifying,yang2021dolg}, R101-DOLG \cite{yang2021dolg} as teacher models, considering their being open-source and decent performance on the landmark image retrieval task. R refers to ResNet~\cite{he2016deep}. We collect the best-performing versions of these pre-trained models from their open-source repository.
For teacher models R101-DELG and R101-DOLG with optional local feature outputs, we only use their global outputs.
Shallower networks ResNet-18/34 (R18/34) are utilized as the student model.
Note that R18 is the default choice unless otherwise stated.

\subsubsection{Influence on teacher models' performance}
The first concern raised by the utilization of whitening is whether it will compromise the performance of teacher models.
Table \ref{tab:eofwt_gldv2} summarizes the performance of each teacher model with and without whitening.
It can be observed that the capacities of teacher models are largely retained after whitening.
The limited influence of whitening on the teacher models' performance provides as a preliminary guarantee for multi-teacher distillation using whitened teachers.

\begin{table}[tbp]
    \begin{center}
    \scalebox{0.71}{
    \begin{tabular}{l*{5}c}
        \toprule
        \multirow{2}{*}{Method} & \multirow{2}{*}{Whitening?} & \multicolumn{2}{c}{$\mathcal{R}$Oxf} & \multicolumn{2}{c}{$\mathcal{R}$Par}\\
        \cmidrule(l){3-4}
        \cmidrule(l){5-6}
        & & M & H & M & H \\
        \midrule
        \multirow{2}{*}{R101-GeM} & \ding{55} & 69.88 & 45.00 & 82.69 & 65.13\\
        & \checkmark & 68.97 & 45.08 & 82.15 & 65.02 \\
        \midrule
        \multirow{2}{*}{R101-AP-GeM} & \ding{55} & 69.17 & 44.08 & 80.44 & 62.08\\
        & \checkmark & 70.37 & 45.93 & 81.02 & 63.29 \\
        \midrule
        \multirow{2}{*}{R101-SOLAR} & \ding{55} & 71.14 & 46.63 & 83.04 & 66.24\\
        & \checkmark & 68.52 & 43.65 & 81.82 & 64.39\\
        \midrule
        \multirow{2}{*}{R101-DELG} & \ding{55} & 84.43 & 66.69 & 91.97 & 82.87\\
        & \checkmark & 85.28 & 68.52 & 92.08 & 83.26\\
        \midrule
        \multirow{2}{*}{R101-DOLG} & \ding{55} & 80.09 & 61.64 & 88.68 & 77.01\\
        & \checkmark & 83.27 & 64.95 & 89.78 & 78.58\\
        \bottomrule
    \end{tabular}
    }
    \end{center}
    \caption{Influence of whitening on the performance of teacher models. The capacities of teacher models are largely retained after whitening. M: Medium, H: Hard.
    }
    \label{tab:eofwt_gldv2}
\end{table}
\begin{table}[tbp]
    \begin{center}
    \scalebox{0.8}{
    \begin{tabular}{l*{5}c}
        \toprule
        \multirow{2}{*}{Strategy} & \multirow{2}{*}{Whitening?} & \multicolumn{2}{c}{$\mathcal{R}$Oxf} & \multicolumn{2}{c}{$\mathcal{R}$Par}\\
        \cmidrule(l){3-4}
        \cmidrule(l){5-6}
        & & M & H & M & H \\
        \midrule
        mean & \multirow{5}{*}{\ding{55}} & 69.62 & 44.01 & 81.62 & 63.64\\
        rand & & 65.68 & 38.89 & 75.78 & 55.09\\
        max-min & & 67.63 & 40.78 & 82.00 & 65.01\\
        max-mean & & 67.09 & 40.07 & 83.66 & 66.93\\
        max-rand & & 71.53 & 46.71 & 80.26 & 62.15\\
        \midrule
        mean & \multirow{5}{*}{\checkmark} & 71.11 & 46.38 & 82.55 & 65.62\\
        rand & & 71.01 & 46.39 & 82.06 & 64.01\\
        max-min & & \textbf{74.67} & \textbf{50.69} & \textbf{84.48} & \textbf{68.34}\\
        max-mean & & 73.90 & 49.53 & 83.42 & 66.55\\
        max-rand & & 72.53 & 48.10 & 82.60 & 65.46\\
        \bottomrule
    \end{tabular}
    }
    \end{center}
    \caption{Performance comparison of different similarity fusion strategies in distillation from R101-GeM, R101-AP-GeM and R101-SOLAR to ResNet-18. 
    The whitening is beneficial for multi-teacher distillation, and it also makes it less sensitive to the selection of fusion strategies. The max-min fusion strategy performs the best with whitening.
    }
    \label{tab:eoffs_gldv2}
\end{table}

\subsubsection{Effectiveness of whitening}
As shown in Table \ref{tab:eoffs_gldv2}, no matter what fusion strategy is adopted, the whitening group can bring consistent performance gains than the non-whitening group, which indicates that whitening teacher models can boost multi-teacher distillation.
Without whitening, the performance shows inconsistency between $\mathcal{R}$Oxf and $\mathcal{R}$Par (\eg max-mean versus max-rand) and unstable between strategies, which is probably due to the direct fusion of incommensurable similarities. In contrast, whitening makes it less sensitive to the selection of fusion strategies and improves on both datasets.

\subsubsection{Empirical studies on fusion strategies}
Among all the strategies, \textit{max-min} performs the best when whitening is adopted. We attribute it to that \textit{max-min} is the most consistent strategy with the objective of visual retrieval where positive pairs are supposed to be embedded closer together and negative pairs further apart. Taking larger positive similarities~(max) and smaller negative similarities~(min) will produce a fused similarity matrix with higher quality than that of an individual teacher: similarities between positive pairs are more distinguished from negative pairs, therefore the overall ranks of positive samples can be improved.
Additionally, \textit{mean} strategy by taking the average of all similarities produces moderate performance. This demonstrates that taking the average is not always an effective strategy for multi-teacher distillation, though it has been the \textit{de facto} choice in most works \cite{ba2014deep,hinton2015distilling,you2017learning}. For the remaining strategies, \ie \textit{max-mean}, \textit{max-rand}, \textit{rand}, when teachers are not whitened, the performance is at most comparable to whitened counterparts. Based on the results, \textit{max-min} strategy will be adopted in later experiments.

\begin{table*}[htbp]
    \begin{center}
    \scalebox{0.75}{
    \begin{tabular}{ll*{10}c}
        \toprule
        \multirow{2}{*}{Method} & \multirow{2}{*}{Teacher models} & \multirow{2}{*}{Params (M)} & \multirow{2}{*}{GFLOPs} & \multicolumn{2}{c}{$\mathcal{R}$Oxf} & \multicolumn{2}{c}{$\mathcal{R}$Par} & \multicolumn{2}{c}{$\mathcal{R}$Oxf+1M} & \multicolumn{2}{c}{$\mathcal{R}$Par+1M}\\
        \cmidrule(l){5-6}
        \cmidrule(l){7-8}
        \cmidrule(l){9-10}
        \cmidrule(l){11-12}
        & & & & M & H & M & H & M & H & M & H \\
        \midrule
        R101-GeM & - & 46.70 & 124 & 68.97 & 45.08 & 82.15 & 65.02 & 55.68 & 29.24 & 61.73 & 35.07 \\
        R101-AP-GeM & - & 46.70 & 124 & 70.37 & 45.93 & 81.02 & 63.29 & 57.48 & 32.01 & 61.26 & 35.59 \\
        R101-SOLAR & - & 56.15 & 139 & 68.52 & 43.65 & 81.82 & 64.39 & 56.24 & 29.65 & 61.53 & 35.12 \\
        \midrule
        \multirow{3}{*}{\textit{\makecell{Single-teacher\\distillation}}} & R101-GeM & \multirow{3}{*}{11.44} & \multirow{3}{*}{28.62} & 70.95 & 46.42 & 81.68 & 64.03 & 55.82 & 28.77 & 61.04 & 33.90 \\
        & R101-AP-GeM & & & 71.41 & 46.46 & 80.64 & 62.82 & 57.61 & 31.17 & 59.67 & 33.57 \\
        & R101-SOLAR & & & 69.78 & 45.12 & 81.78 & 63.76 & 55.87 & 28.84 & 60.63 & 34.04 \\
        \midrule
        \multirow{3}{*}{\textit{\makecell{Double-teacher\\distillation}}} & R101-GeM, -AP-GeM & \multirow{3}{*}{11.44} & \multirow{3}{*}{28.62} & 73.90 & 49.81 & 84.09 & 67.91 & 61.12 & 33.72 & 64.44 & 38.89 \\
        & R101-GeM, -SOLAR & & & 70.79 & 46.95 & 82.48 & 64.93 & 56.20 & 29.41 & 62.57 & 36.13 \\
        & R101-AP-GeM, -SOLAR & & & \textbf{74.71} & 50.21 & 83.59 & 67.01 & 60.93 & 33.01 & 64.28 & 39.53 \\
        \midrule
        \textit{\makecell{Triple-teacher\\distillation}} & R101-GeM, -AP-GeM, -SOLAR & 11.44 & 28.62 & 74.67 & \textbf{50.69} & \textbf{84.48} & \textbf{68.34} & \textbf{61.74} & \textbf{34.49} & \textbf{64.82} & \textbf{39.63} \\
        \bottomrule
    \end{tabular}
    }
    \end{center}
    \caption{Comparison of original teacher models and distillation models using varying teacher combinations in terms of model complexity and performance. All the distillation variants utilize R18 as the student model.
    Multi-teacher distillation methods not only reduce the model complexity but also achieve further performance gain.
    }
    \label{tab:eofmt_gldv2}
\end{table*}

\subsubsection{Single-teacher distillation vs. Multi-teacher distillation}
Table \ref{tab:eofmt_gldv2} summarizes the performance and complexity of original teacher models and distillation models using varying teacher combinations.
The first group reports the performance of whitened teacher models.
As shown in the table, single-teacher distillation could significantly reduce the model complexity during the inference, but their performance is only comparable to the corresponding teacher models.
By contrast, multi-teacher distillation methods including double-teacher and triple-teacher distillation not only reduce the model complexity but also achieve further performance gain.
We attribute the performance gain to the complementarity of multiple teacher models and the effectiveness of our proposed multi-teacher distillation framework.
Besides, the triple-teacher distillation almost surpassing the double-teacher distillation demonstrates the incremental advantage of fusing more teacher models.

It is possible to improve even further by introducing more teacher models. But we can also imagine a performance saturation with endless addition of teachers, due to the student model's limited capacity, less complementarity between teachers, etc. Using more teacher models also means larger training cost. It is a trade-off in practice and we recommend determining the number of teachers based on the expected service time to make the additional training cost more amortized by the improved inference speed.

\subsubsection{Comparison to baseline approaches}

\begin{table}[tbp]
    \begin{center}
    \scalebox{0.65}{
    \begin{tabular}{l*{6}c}
        \toprule
        \multirow{2}{*}{Method} & \multirow{2}{*}{Params (M)} & \multirow{2}{*}{GFLOPs} & \multicolumn{2}{c}{$\mathcal{R}$Oxf} & \multicolumn{2}{c}{$\mathcal{R}$Par}\\
        \cmidrule(l){4-5}
        \cmidrule(l){6-7}
        & & & M & H & M & H\\
        \midrule
        Ensemble Mean (EM) & 149.55 & 387 & 71.14 & 46.67 & 83.38 & 66.41\\
        Embedding Distillation (ED) & 11.44 & 28.62 & 68.19 & 42.29 & 80.47 & 61.73\\
        Contrastive Learning (CL) & 11.44 & 28.62 & 65.72 & 40.25 & 81.78 & 63.69\\
        Triple-teacher distillation & 11.44 & 28.62 & \textbf{74.67} & \textbf{50.69} & \textbf{84.48} & \textbf{68.34}\\
        \bottomrule
    \end{tabular}
    }
    \end{center}
    \caption{Comparison to common baseline approaches.
    Our proposed multi-teacher distillation framework could be an alternative to model ensemble learning, and much more lightweight and effective for landmark image retrieval.
    }
    \label{tab:baselines_gldv2}
\end{table}
In order to further verify the effectiveness of multi-teacher distillation, we compare it to three related baseline approaches, including Ensemble Mean (EM), Embedding Distillation (ED), and Contrastive Learning (CL).
EM is a classical ensemble method, which averages the similarity scores produced by teacher models during inference. 
ED is a feature-based distillation method, which jointly minimizes the Euclidean distances between output embeddings of teacher models and the student model. 
CL is a representation learning method widely used in recent works \cite{chen2021empirical}, which is equivalent to replacing $\mathcal{Q}$ in Eq. \ref{eq:loss} with an identity matrix.

As in Table \ref{tab:baselines_gldv2}, our proposed triple-teacher distillation consistently outperforms the counterparts with a clear margin.
Among them, EM fusing knowledge from multiple models performs the best, but it brings heavy computation overhead as every ensembled model must be forwarded to calculate its similarity output. 
By contrast, our proposed multi-teacher distillation framework only needs to forward a lightweight student once during inference.
Additionally, the worse performance of ED than ours demonstrates that relation-based distillation is more suitable for multi-teacher distillation.
\begin{figure}[tbp]
  \centering
  \begin{subfigure}{0.42\columnwidth}
    \includegraphics[width=\linewidth]{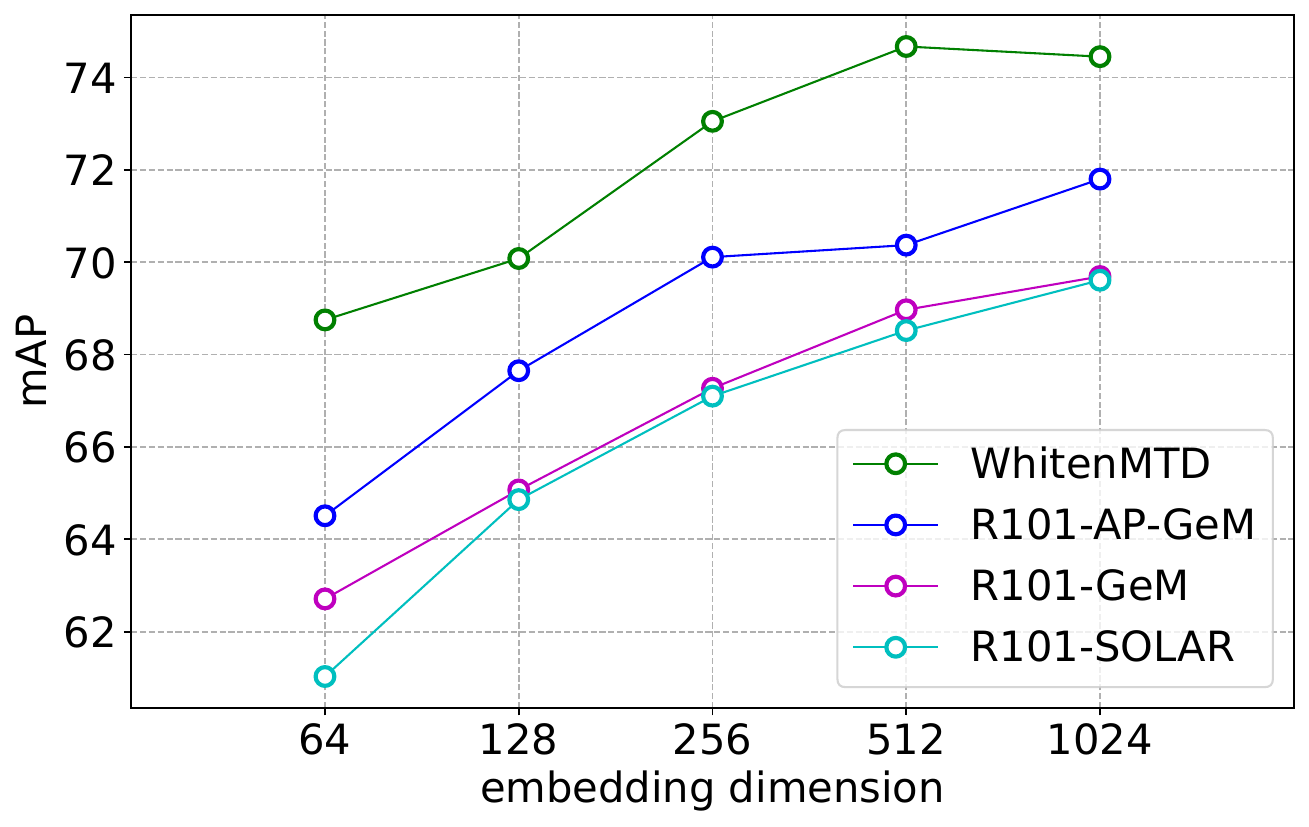}
    \caption{$\mathcal{R}$Oxf-M}
    \label{fig:eof_stu_dim_a}
  \end{subfigure}
  \begin{subfigure}{0.42\columnwidth}
    \includegraphics[width=\linewidth]{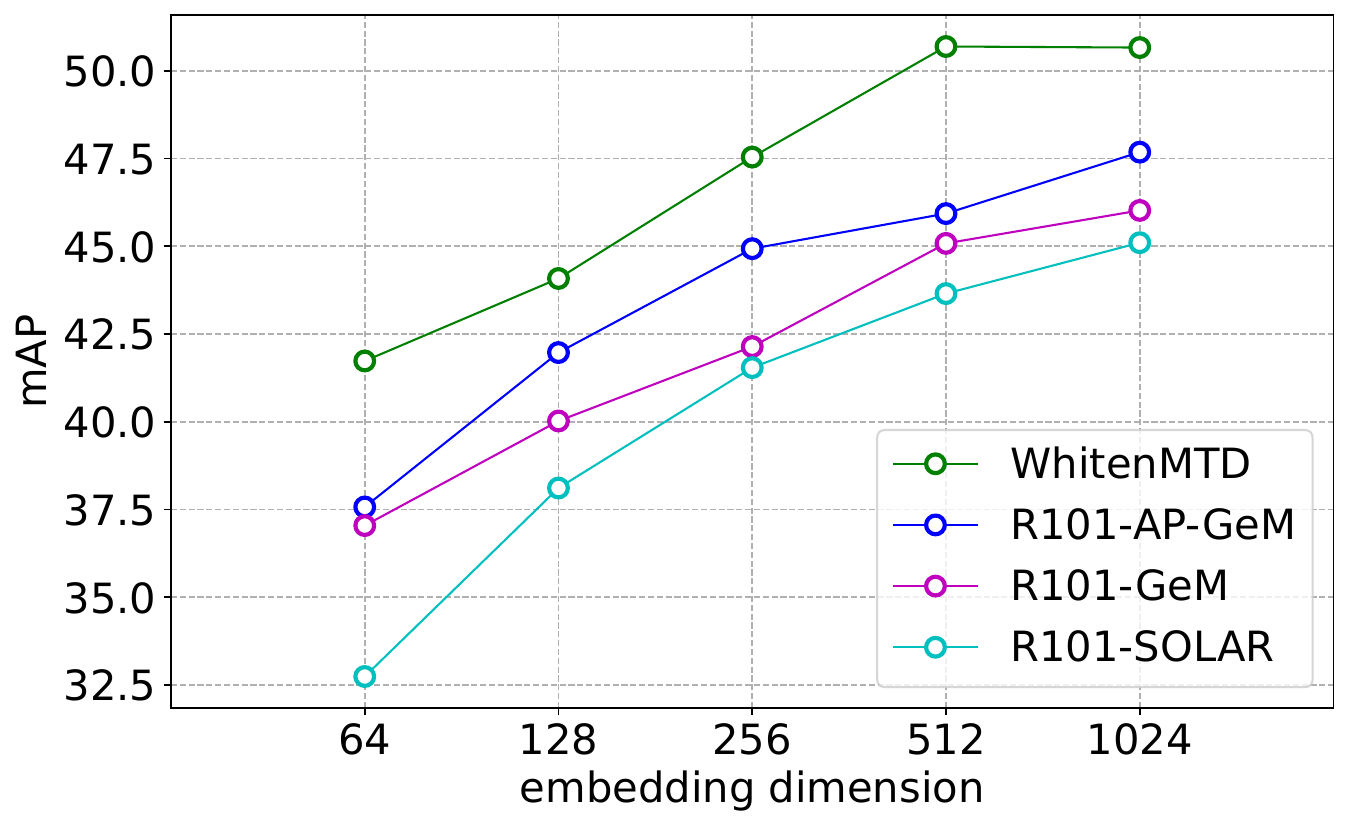}
    \caption{$\mathcal{R}$Oxf-H}
    \label{fig:eof_stu_dim_b}
  \end{subfigure}
  \hfill
  \begin{subfigure}{0.42\columnwidth}
    \includegraphics[width=\linewidth]{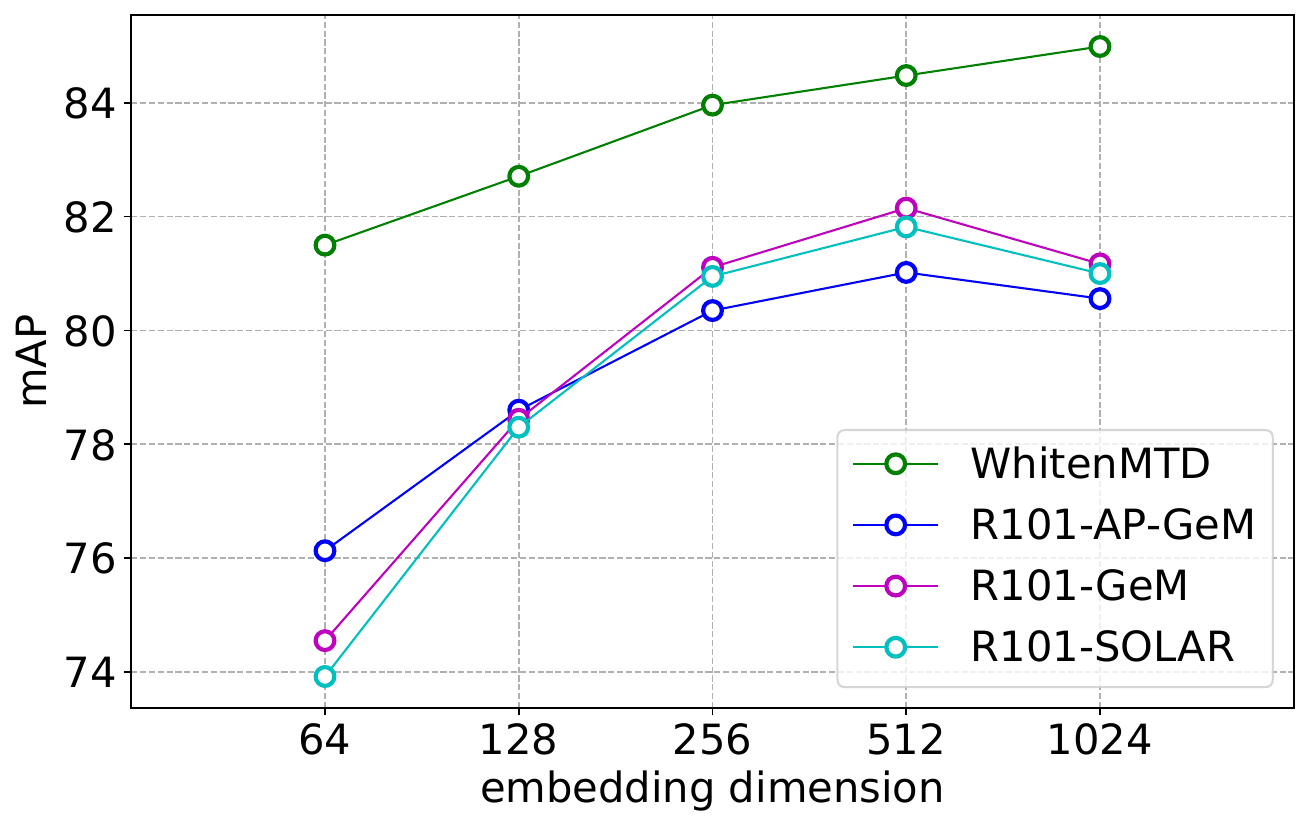}
    \caption{$\mathcal{R}$Par-M}
    \label{fig:eof_stu_dim_c}
  \end{subfigure}
  \begin{subfigure}{0.42\columnwidth}
    \includegraphics[width=\linewidth]{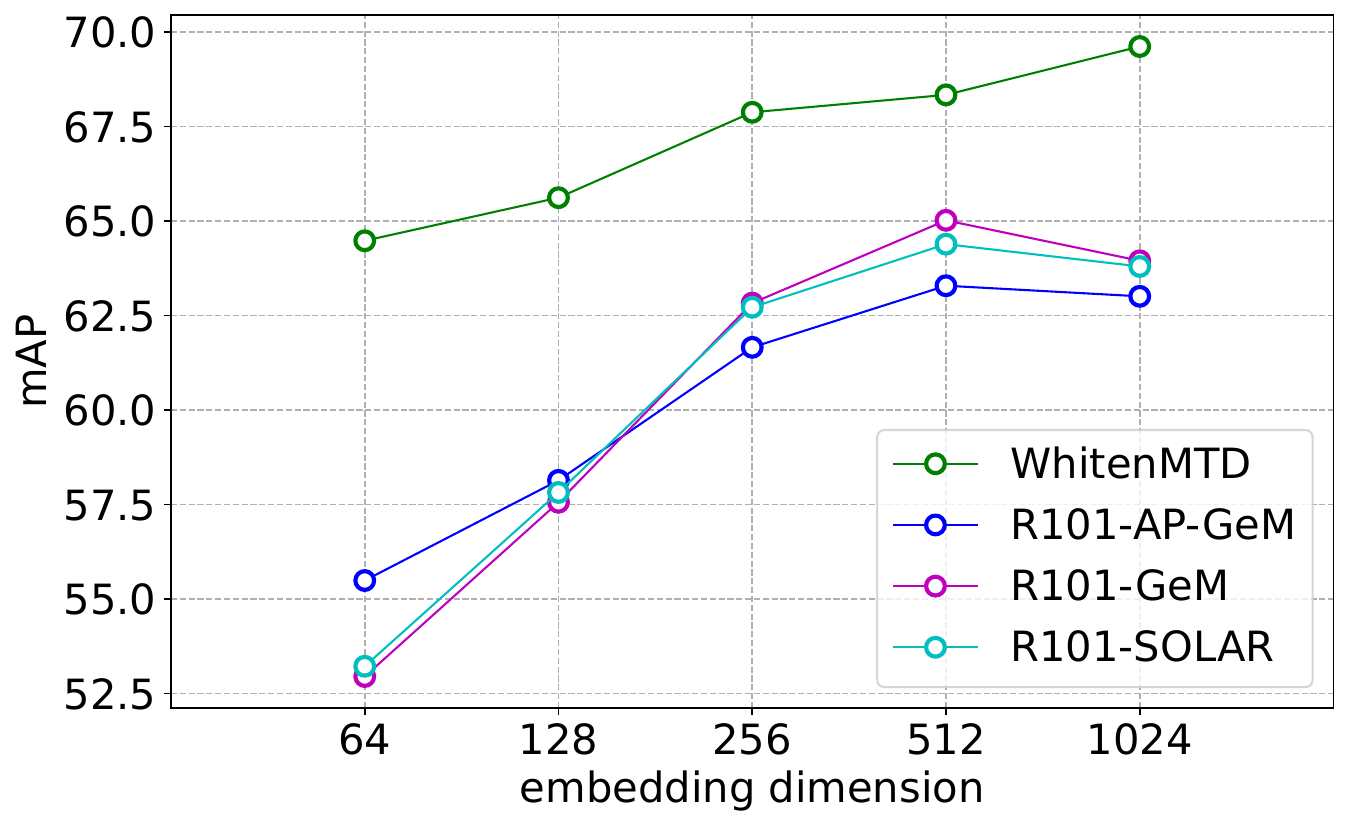}
    \caption{$\mathcal{R}$Par-H}
    \label{fig:eof_stu_dim_d}
  \end{subfigure}
  \caption{
  Performance of varying output dimension on $\mathcal{R}$Oxf and $\mathcal{R}$Par datasets. The output dimensions of R101-GeM, R101-AP-GeM and R101-SOLAR are changed by whitening. Our proposed WhitenMTD consistently outperforms them with clear margins.
  }
  \label{fig:eof_stu_dim}
\end{figure}
\begin{table*}[htbp]
    \begin{center}
    \scalebox{0.8}{
    \begin{tabular}{l*{10}c}
        \toprule
        \multirow{2}{*}{Method} & \multirow{2}{*}{Params (M)} & \multirow{2}{*}{GFLOPs} & \multicolumn{2}{c}{$\mathcal{R}$Oxf} & \multicolumn{2}{c}{$\mathcal{R}$Par} & \multicolumn{2}{c}{$\mathcal{R}$Oxf+1M} & \multicolumn{2}{c}{$\mathcal{R}$Par+1M}\\
        \cmidrule(l){4-5}
        \cmidrule(l){6-7}
        \cmidrule(l){8-9}
        \cmidrule(l){10-11}
        & & & M & H & M & H & M & H & M & H \\
        \midrule
        \multicolumn{11}{l}{\textit{(A) Local feature aggregation}}\\
        R50-DELF-ASMK+SP (\citeauthor{noh2017large,radenovic2018revisiting}) & 9.07 & 53.82 & 67.80 & 43.10 & 76.90 & 55.40 & 53.80 & 31.20 & 57.30 & 26.40\\
        R50-DELF-R-ASMK+SP (\citeauthor{teichmann2019detect}) & 9.07 & 53.82 & 76.00 & 52.40 & 80.20 & 58.60 & 64.00 & 38.10 & 59.70 & 29.40\\
        R50-HOW-ASMK (\citeauthor{tolias2020learning}) & 8.67 & 52.61 & 79.40 & 56.90 & 81.60 & 62.40 & 65.80 & 38.90 & 61.80 & 33.70\\
        \midrule
        \multicolumn{11}{l}{\textit{(B) Global features + local features}}\\
        R101-GeM$\uparrow$+DSM (\citeauthor{simeoni2019local}) & 42.50 & 124 & 65.30 & 39.20 & 77.40 & 56.20 & 47.60 & 23.20 & 52.80 & 25.00\\
        R101-DELG (\citeauthor{cao2020unifying,yang2021dolg}) & 44.08 & 125 & 81.20 & 64.00 & 87.20 & 72.80 & 69.10 & 47.50 & 71.50 & 48.70 \\
        R50-DELG+RRT (\citeauthor{tan2021instance}) & 25.08 & 66.55 & 78.10 & 60.20 & 86.70 & 75.10 & 67.00 & 44.10 & 69.80 & 49.40\\
        \midrule
        \multicolumn{11}{l}{\textit{(C) Global features}}\\
        R101-GeM (\citeauthor{radenovic2018fine}) & 46.70 & 124 & 68.43 & 44.41 & 82.19 & 65.05 & 55.25 & 30.04 & 61.77 & 35.14 \\
        R101-AP-GeM (\citeauthor{revaud2019learning}) & 46.70 & 124 & 70.00 & 45.60 & 80.91 & 63.00 & 57.36 & 31.89 & 61.09 & 35.27 \\
        R101-SOLAR (\citeauthor{ng2020solar}) & 56.15 & 139 & 69.90 & 47.90 & 81.60 & 64.50 & 53.50 & 29.90 & 59.20 & 33.40\\
        R101-DELG (\citeauthor{cao2020unifying,yang2021dolg}) & 43.55 & 124 & 76.30 & 55.60 & 86.60 & 72.40 & 63.70 & 37.50 & 70.60 & 46.90 \\
        R101-DOLG (\citeauthor{yang2021dolg}) & 46.70 & 127 & 81.50 & 61.10 & 91.02 & 80.30 & 77.43 & 54.81 & 83.29 & 66.69 \\
        R50-CVNet-Global+Rerank (\citeauthor{lee2022correlation}) & 35.20 & 64.41 & \textbf{87.90} & 75.60 & 90.50 & 80.20 & 80.70 & 65.10 & 82.40 & 67.30 \\
        R101-CVNet-Global+Rerank (\citeauthor{lee2022correlation}) & 54.20 & 123 & 87.20 & \textbf{75.90} & 91.20 & 81.10 & \textbf{81.90} & \textbf{67.40} & 83.80 & 69.30 \\
        DToP-R50+ViT-B (\citeauthor{song2023boosting}) & - & - & 84.40 & 64.80 & 92.30 & \textbf{84.60} & 78.90 & 57.10 & \textbf{85.40} & \textbf{71.20} \\
        \textbf{R18-\modelname~(ours)} & 11.44 & 28.62 & 81.71 & 59.73 & 90.67 & 80.15 & 74.05 & 46.48 & 77.24 & 57.90 \\
        \textbf{R34-\modelname~(ours)} & 21.55 & 57.71 & 83.05 & 64.16 & \textbf{92.33} & 83.20 & 76.88 & 52.65 & 80.57 & 63.26 \\
        \bottomrule
    \end{tabular}
    }
    \end{center}
    \vspace{-3mm}
    \caption{Comparison to state-of-the-art on $\mathcal{R}$Oxf and $\mathcal{R}$Par datasets. All scores are from their original reports, except R101-DELG which is re-implemented by \cite{yang2021dolg} and trained on GLDv2 showing better performance than the original. The best scores are marked in bold. Our models of using only global features are distilled from R101-DOLG and R101-DELG, which demonstrates a good balance of retrieval performance and efficiency.
    }
    \label{tab:sota_gldv2}
    \vspace{-3mm}
\end{table*}
\begin{table}[htbp]
    \begin{center}
    \scalebox{0.65}{
        \begin{tabular}{l*{4}c}
        \toprule
        Method & Params (M) & GFLOPs & mAP@100 & mAP \\
        \midrule
        VGG16-CNNL (\citeauthor{kordopatis2017near1}) & 134 & 15.47 & 61.04 & 55.55 \\
        VGG16-CNNV (\citeauthor{kordopatis2017near1}) & 134 & 15.47 & 25.10 & 19.09 \\
        VGG16-CTE (\citeauthor{revaud2013event}) & 134 & 15.47 & - & 50.97 \\
        VGG16-DML (\citeauthor{kordopatis2017near2}) & 139 & 15.47 & 81.27 & 78.47 \\
        R50-VRL (\citeauthor{he2022learn}) & 23.5 & 4.14 & 86.00 & - \\
        R50-DnS (\citeauthor{kordopatis2022dns}) & 27.55 & 4.13 & - & \textbf{90.20} \\
        \midrule
        R50-MoCoV3 (\citeauthor{chen2021empirical}) & 23.5 & 4.14 & 87.31 & 85.47 \\
        R50-BarlowTwins (\citeauthor{zbontar2021barlow}) & 23.5 & 4.14 & 87.22 & 84.80 \\
        \midrule
        \textbf{R18-\modelname~(ours)} & 11.2 & 1.83 & 88.62 & 86.82 \\
        \textbf{R34-\modelname~(ours)} & 21.3 & 3.68 & \textbf{88.84} & 86.78 \\
        \bottomrule
        \end{tabular}
    }
    \end{center}
    \vspace{-3mm}
    \caption{Comparison to state-of-the-art on SVD. MoCoV3 and BarlowTwins are the teacher models. Whiten-MTD with a more lightweight backbone achieves comparable performance.}
    \label{tab:sota_svd}
    \vspace{-4mm}
\end{table}

\subsubsection{Embedding dimension of student model}
Not only does the computation overhead account for efficient visual retrieval, but also the embedding dimension is a crucial consideration in practice. A smaller embedding dimension of retrieval models is preferred as it can greatly reduce the storage and computation cost. Therefore, we study the influence of the output embedding dimension on the final retrieval performance. 
We compare to R101-GeM, R101-AP-GeM and R101-SOLAR, which are jointly used as the teacher models in our Whiten-MTD.
As depicted in Figure \ref{fig:eof_stu_dim}, the student model consistently outperforms each teacher model with clear margins.
The results highlight the effectiveness of our multi-teacher distillation framework.

\subsubsection{Comparison to state-of-the-art}
Table \ref{tab:sota_gldv2} summarizes the performance comparison on $\mathcal{R}$Oxf and $\mathcal{R}$Par, where the state-of-the-art methods are categorized into three groups depending on the type of features used. Note that we only utilize the global features for evaluation, hence our approach belongs to the third group.
To make it more comparable to state-of-the-art methods, we choose two best-performing methods R101-DELG and R101-DOLG as teacher models.

Compared to existing methods based on the deep R50/R101 as the backbone, our \modelname~achieves comparable performance even if R18 is utilized as the student model. Using a larger R34 as the student model brings further performance boost, which achieves the best result on $\mathcal{R}$Par-M and competitive results otherwise.
Even compared against reranking methods, our model still obtains competitive results with fewer parameters and GFLOPs.
It is worth pointing out that models in the first group have fewer parameters than ours due to the last  block being removed from R50, but their performance is much worse. 
What's more, when distilling from the two best-performing R101-DELG and R101-DOLG, the students are slightly worse than the teachers (see Table \ref{tab:eofwt_gldv2}) but instead more efficient.
The results demonstrate that our \modelname~accomplishes a good balance of retrieval performance and efficiency.

\subsection{Near-duplicate video retrieval}
\subsubsection{Experimental setup}
We utilize Short Video Dataset (SVD)~\cite{jiang2019svd}, considering it is the latest and the largest benchmark dataset for near-duplicate video retrieval. We collect two self-supervised pre-trained models as teachers, \ie R50-MoCoV3 \cite{chen2021empirical}, R50-BarlowTwins \cite{zbontar2021barlow}.
For the student model, we also use R18 and R34. To make a fair comparison with state-of-the-art works \cite{he2022learn,kordopatis2022dns}, we report both mAP@100 and mAP.

\subsubsection{Results}
The results on the SVD dataset are shown in Table \ref{tab:sota_svd}.
With a more lightweight backbone, our model achieves comparable performance with mAP@100 of 88.84 and mAP of 86.82. State-of-the-art method 
\cite{kordopatis2022dns} relies on computationally expensive spatial-temporal feature extraction and reranking to achieve a better result while we only perform a simple forward of R18 or R34. The results demonstrate the effectiveness of our multi-teacher distillation framework for video retrieval.

\section{Conclusion}
\label{sec:conclusion}
In this paper we contribute a multi-teacher distillation method for efficient visual retrieval. We propose a simple and effective multi-teacher distillation framework \modelname. We also investigate five heuristic fusion strategies and make a detailed analysis of their adaptations. For effective multi-teacher distillation whitening is needed before applying any fusion strategies to eliminate the similarity distribution discrepancy of teacher models. Extensive experiments demonstrate that the student models obtained using our method are computationally efficient, with comparable performance to state-of-the-art methods based on heavy networks on both instance image retrieval and video retrieval. In the future, we will explore our proposed multi-teacher distillation in cross-modal retrieval~\cite{fang2023you,dong2022dual,zheng2023progressive,fang2023you}.

\appendix
\section{Acknowledgments}
This work is sponsored by the National Key Research, Development Program of China under No. 2022YFB3102100, Pioneer and Leading Goose R\&D Program of Zhejiang (No. 2023C01212), Young Elite Scientists Sponsorship Program by CAST (No. 2022QNRC001), National Natural Science Foundation of China (No. 61976188, No. 62372402), CCF-AFSG Research Fund, R\&D Program of DCI Technology Application Joint Laboratory.

\bibliography{ref}

\end{document}


\maketitle

Due to the limited space, we here report more experimental results and technical details which are not included in the main paper:
\begin{itemize}
\item Further discussion on whitening.
\item Performance and complexity comparison to asymmetric image retrieval approaches \cite{wu2022contextual,budnik2021asymmetric,duggal2021compatibility} on $\mathcal{R}$Oxf and $\mathcal{R}$Par.
\item Influence of temperature factor $\tau$ used in distillation.
\item Analysis of training statistics including similarity scores and mean reciprocal rank of teacher models to provide a better understanding of the intuition of our method.
\item Full implementation details.
\end{itemize}

\section{Further discussion on whitening}

In this section we detail the PCA-whitening process, provide some guidance on the choice of the dimensionality $n_c$ after whitening and illustrate the cosine similarity distribution after whitening both therectically and statistically.

Given $m$-dimensional input random vector $\mathbf{x}\in\mathbb{R}^{m\times 1}$, a teacher model $\psi$ represents it as an $n_t$-dimensional feature vector $\mathbf{z}=\psi(\mathbf{x})\in \mathbb{R}^{n_t\times 1}$. Note that we omit the teacher index $k$ here for simplicity. $\mathbf{z}$ is first converted to have zero mean:
\begin{equation}
\bar{\mathbf{z}}=\mathbf{z}-\mathbb{E}[\mathbf{z}].
\end{equation}
$\bar{\mathbf{z}}$ has a real symmetric positive semi-definite covariance matrix $\mathbf{\Sigma}\in\mathbb{R}^{n_t\times n_t}$. Whitening linearly transforms $\bar{\mathbf{z}}$ to a new random vector $\mathbf{u}\in\mathbb{R}^{n_c}$ with unit diagonal covariance:
\begin{gather}
\mathbf{u}=\mathbf{W}\bar{\mathbf{z}} \notag\\
\mathrm{s.t.} \mathbf{W}^\top\mathbf{W}=\mathbf{\Sigma}^{-1}.
\end{gather}
Although there are theoretically in fact infinite possible matrices $\mathbf{W}$ that can be used to perform whitening, \eg PCA-whitening, ZCA-whitening, Cholesky whitening \cite{kessy2018optimal}, in our method we adopt PCA-whitening. PCA-whitening starts by the eigenvalue decomposition of the covariance matrix $\mathbf{\Sigma}$:
\begin{equation}
\mathbf{\Sigma}=\mathbf{U}\mathrm{diag}(\mathbf{\Lambda})\mathbf{U^\top},
\end{equation}
where $\mathbf{U}\in\mathbb{R}^{n_t\times n_t}$ is the an orthogonal matrix composed of eigenvectors and $\mathbf{\Lambda}\in\mathbb{R}^{n_t}$ consists of $n_t$ eigenvalues. Assume $n_t$ eigenvalues $[\lambda_1,\lambda_2,...,\lambda_{n_t}]$ are arranged in descending orders. When $\forall i\in[1,n_t],\lambda_i>0$, the whitening transformation matrix $\mathbf{W}$ is:
\begin{equation}
    \mathbf{W}=\mathrm{diag}(\mathbf{\Lambda})^{-1/2}\mathbf{U^\top}.
\end{equation}

In practice, however, the eigenvalues are not necessarily to be all positive. There can be $s$ almost zero eigenvalues, which indicates the representation $\bar{\mathbf{z}}$ actually distributes in a $s$-dimensional subspace of the whole space. Table \ref{tab:num_of_sig_eigen} shows the specific number of significant components of each teacher models trained on GLDv2. It can be seen that R101-AP-GeM and R101-DOLG do not have full-rank representation space. Recall that we have demonstrated in the main paper that whitened and $l_2$-normalized representations distribute uniformly on the unit sphere surface and thus have a deterministic cosine similarity density function, which eliminates the similarity distribution discrepancy among different teacher models. An additional condition must be fulfilled when considering the sparsity of the representation space:

\textbf{\textit{If $n_c\leq n_t-s$, the similarity distribution of different teacher models will be definitely aligned.}}

\begin{table}[tbp]
    \begin{center}
    \scalebox{0.8}{
    \begin{tabular}{lccc}
        \toprule
        Teacher model & $n_t$ & $s$ & $n_t-s$\\
        \midrule
        R101-GeM & 2048 & 0 & 2048\\
        R101-AP-GeM & 2048 & 86 & 1962\\
        R101-SOLAR & 2048 & 0 & 2048\\
        R101-DELG & 512 & 0 & 512\\
        R101-DOLG & 1024 & 611 & 413\\
        \bottomrule
    \end{tabular}
    }
    \end{center}
    \caption{The output dimensions of teacher models and the number of significant eigenvalues ($>10^{-5}$).
    }
    \label{tab:num_of_sig_eigen}
\end{table}
\begin{figure}[tb]
  \centering
  \begin{subfigure}{0.3\linewidth}
    \includegraphics[width=\linewidth]{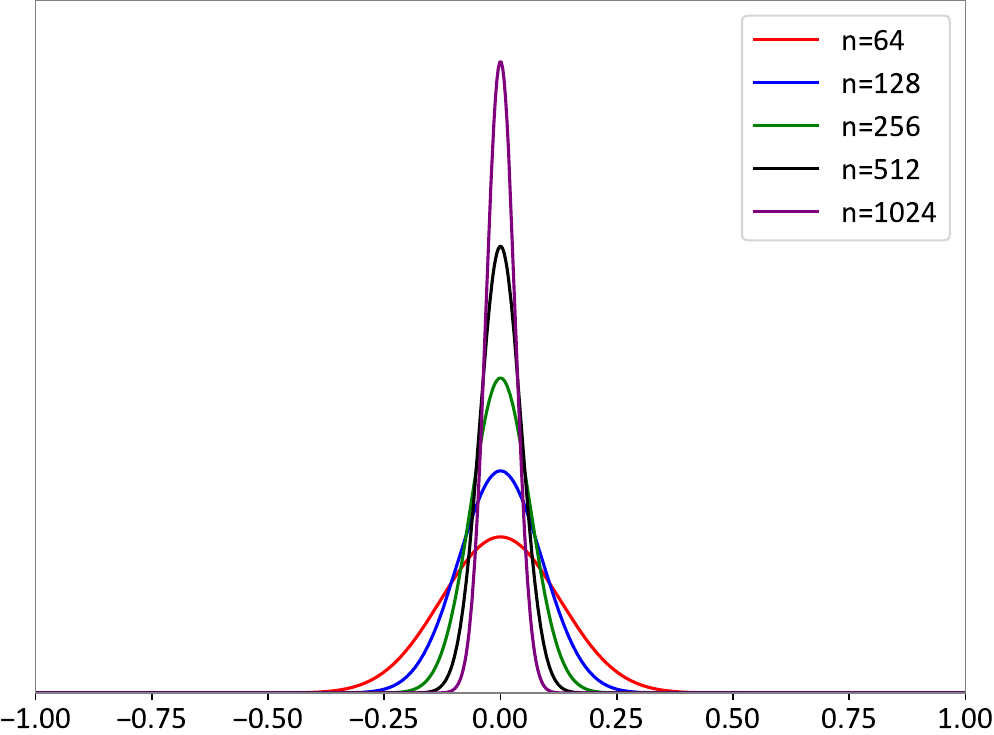}
    \caption{Theoretical}
    \label{fig:sim_whit_dim_theoretical}
  \end{subfigure}
  \hfill
  \begin{subfigure}{0.3\linewidth}
    \includegraphics[width=\linewidth]{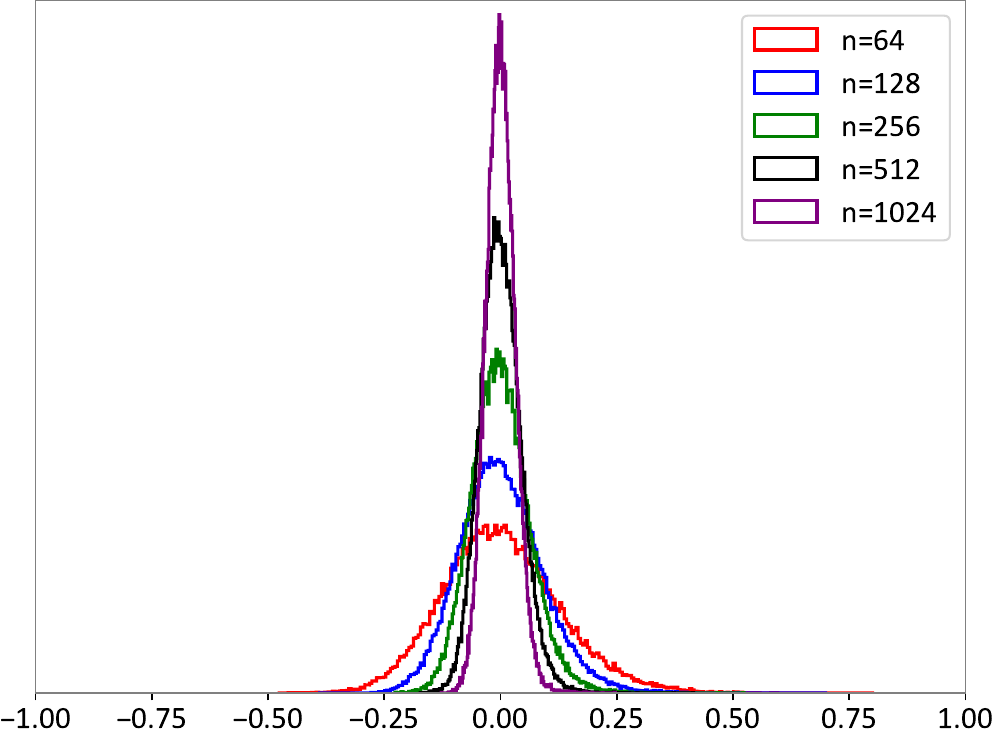}
    \caption{R101-GeM}
    \label{fig:sim_whit_dim_rg}
  \end{subfigure}
  \hfill
  \begin{subfigure}{0.3\linewidth}
    \includegraphics[width=\linewidth]{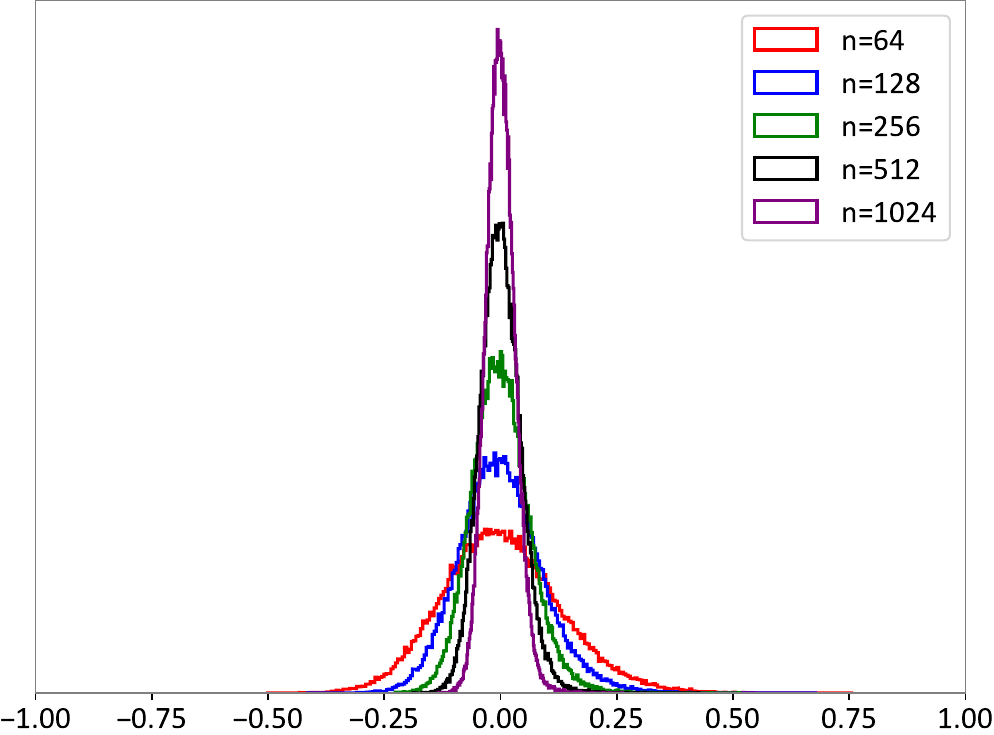}
    \caption{R101-AP-GeM}
    \label{fig:sim_whit_dim_rag}
  \end{subfigure}
  \hfill
  \begin{subfigure}{0.3\linewidth}
    \includegraphics[width=\linewidth]{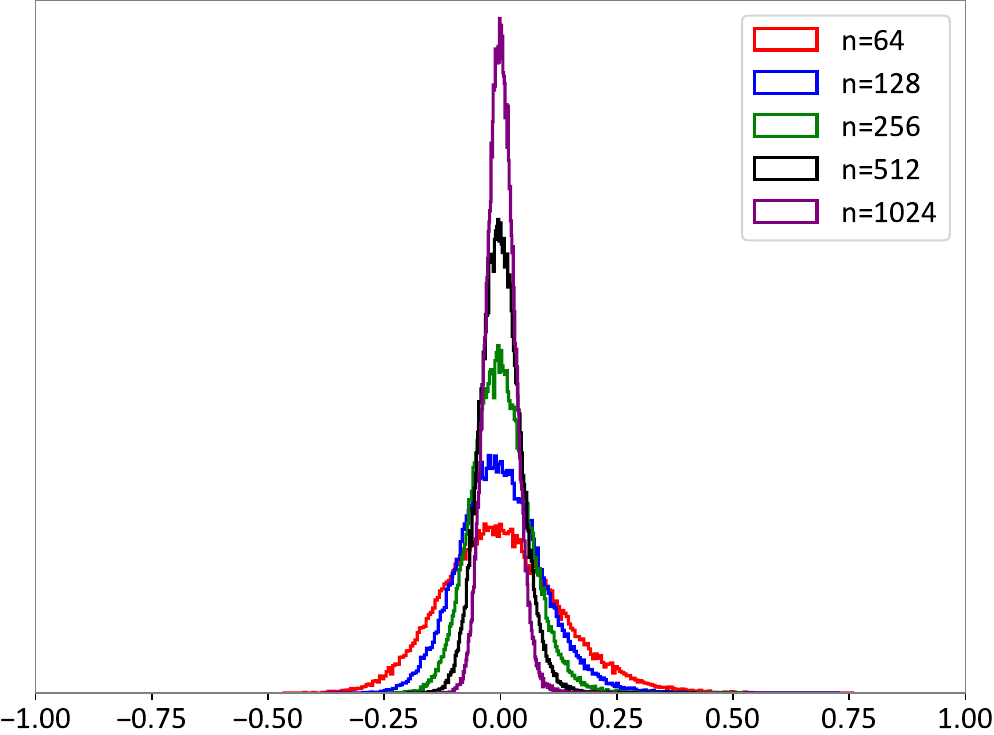}
    \caption{R101-SOLAR}
    \label{fig:sim_whit_dim_rs}
  \end{subfigure}
  \hfill
  \begin{subfigure}{0.3\linewidth}
    \includegraphics[width=\linewidth]{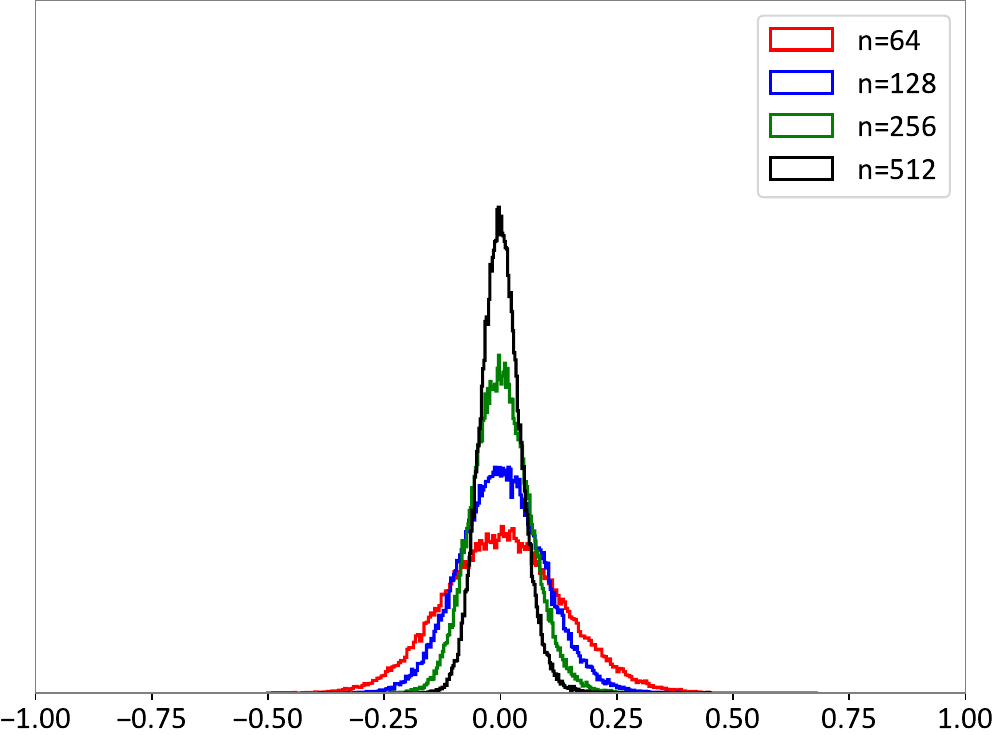}
    \caption{R101-DELG}
    \label{fig:sim_whit_dim_re}
  \end{subfigure}
  \hfill
  \begin{subfigure}{0.3\linewidth}
    \includegraphics[width=\linewidth]{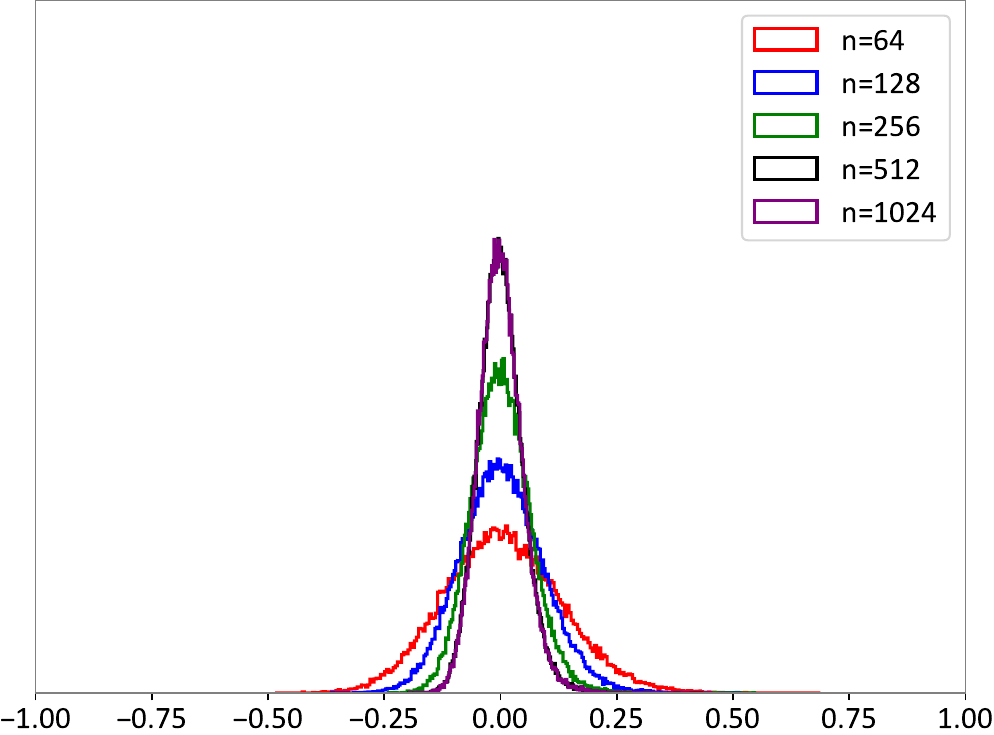}
    \caption{R101-DOLG}
    \label{fig:sim_whit_dim_ro}
  \end{subfigure}
  \caption{Theoretical cosine similarity distribution (a) and statistical results of teacher models (b-f) under different output dimensions after whitening ($n_c$). The statistical results match the theoretical results in all dimensions. Similarity distribution under different output dimensions are of different shapes and higher dimension induces more concentrated similarity distribution. Note that the two curves for $n=512$ and $n=1024$ of model R101-DOLG overlaps because there are only 413 significant components (Table \ref{tab:num_of_sig_eigen}) in its original 1024-dimensional space. This indicates that teacher models' output dimension should be same to perform effective multi-teacher distillation.
  }
  \label{fig:sim_whit_dim}
\end{figure}


Figure \ref{fig:sim_whit_dim} shows that theoretical cosine similarity distribution according to $f(\theta) = \frac{1}{\sqrt{\pi}}\cdot\frac{\Gamma(\frac{n}{2})}{\Gamma(\frac{n-1}{2})}\cdot(\sin\theta)^{n-2},\theta\in\left[ 0,\pi\right]$ and the statistical distribution of teacher models. First, the similarity distributions of all teacher models after whitening can be aligned under the same $n_c$ and they totally match the theoretical results. Second, it can be observed that whitened distributions of different dimensions are of varied shapes but distributions of different models under the same dimension are the same. Based on this observation, we conclude that $n_c$ must be same for all teacher models. In experiments we found that $n_c=512$ is able to produce good results. Note that for teacher R101-DOLG, its similarity curves of $n=512$ and $n=1024$ overlap because there are only 413 significant components (see Table \ref{tab:num_of_sig_eigen}) in its original $1024$-dimensional representation space. The whitened and $l_2$-normalized representations of R101-DOLG actually distribute on a $413$-dimensional unit sphere. Although it is smaller than $n_c=512$, the similarity distribution of R101-DOLG is approximately aligned to other teachers and it still works well in distillation. We leave the representation sparsity problem as a future study.

\section{Influence of temperature}
\label{sec:temp}
The hyperparameter $\tau$ used in distillation usually has considerable affect on the performance. Therefore we study its effect by a simple grid search over teachers' and student temperatures. The teacher models and student may use different temperatures as in previous work \cite{wu2022contextual}. We denote the temperature used by teachers $\tau_t$ and student $\tau_s$. Table \ref{tab:temperature} provides results on different combinations of temperature. Our default setting $\tau_t=\tau_s=0.05$ produces overall the best results. The worst case occurs when $\tau_t$ and $\tau_s$ have too large difference.

\begin{table}[t]
    \begin{center}
    \scalebox{0.8}{
    \begin{tabular}{l*{5}c}
        \toprule
        Teacher $\tau_t$ & Student $\tau_s$ & $\mathcal{R}$Oxf-M & $\mathcal{R}$Oxf-H & $\mathcal{R}$Par-M & $\mathcal{R}$Par-H\\
        \midrule
        \textbf{0.05} & \textbf{0.05} & 74.67 & 50.69 & 84.48 & 68.34\\
        0.01 & 0.01 & 72.55 & 47.25 & 83.85 & 67.49\\
        0.1 & 0.1 & 72.79 & 48.40 & 81.89 & 64.59\\
        0.01 & 0.05 & 72.67 & 48.02 & 84.78 & 68.77\\
        0.01 & 0.1 & 71.57 & 45.27 & 83.27 & 67.17\\
        0.05 & 0.01 & 74.21 & 50.03 & 84.26 & 68.50\\
        0.05 & 0.1 & 74.12 & 49.21 & 84.36 & 68.39\\
        0.1 & 0.01 & 70.85 & 45.98 & 81.72 & 64.53\\
        0.1 & 0.05 & 71.12 & 45.51 & 81.77 & 63.88\\
        \bottomrule
    \end{tabular}
    }
    \end{center}
    \caption{Influence of temperature $\tau$. The default setting is in bold font.
    }
    \label{tab:temperature}
\end{table}

\section{Comparison to asymmetric image retrieval}
\label{sec:asym}
\begin{table*}[tbp]
    \begin{center}
    \scalebox{0.8}{
    \begin{tabular}{l*{12}c}
        \toprule
        \multirow{2}{*}{Method} & \multicolumn{2}{c}{Query model} & \multicolumn{2}{c}{Gallery model} & \multicolumn{2}{c}{$\mathcal{R}$Oxf} & \multicolumn{2}{c}{$\mathcal{R}$Par} & \multicolumn{2}{c}{$\mathcal{R}$Oxf+1M} & \multicolumn{2}{c}{$\mathcal{R}$Par+1M}\\
        \cmidrule(l){2-3}
        \cmidrule(l){4-5}
        \cmidrule(l){6-7}
        \cmidrule(l){8-9}
        \cmidrule(l){10-11}
        \cmidrule(l){12-13}
        & \makecell{Params\\(M)} & GFLOPs &\makecell{Params\\(M)} & GFLOPs & M & H & M & H & M & H & M & H \\
        \midrule
        MobileNetV2-R101-Reg & 4.8 & 0.3 & 43.5 & 7.85 & 72.75 & 53.07 & 85.81 & 69.96 & 56.03 & 32.21 & 65.23 & 39.29 \\
        EfficientNetB3-R101-Reg & 13.84 & 16.14 & 42.50 & 124 & 74.60 & 53.41 & 86.09 & 72.21 & 59.88 & 33.31 & 67.69 & 42.63 \\
        MobileNetV2-R101-Context & 4.8 & 0.3 & 43.5 & 7.85 & 76.01 & 57.61 & 87.55 & 74.82 & 58.42 & 36.58 & 69.24 & 45.67 \\
        EfficientNetB3-R101-DELG-Context & 11.48 & 16.13 & 43.55 & 124 & 77.44 & 58.97 & 87.94 & 75.68 & 63.21 & 38.20 & 73.37 & 50.09 \\
        MobileNetV2-R101-HVS & 4.8 & 0.3 & 43.5 & 7.85 & 74.39 & 54.68 & 86.86 & 72.42 & 58.24 & 34.77 & 67.44 & 43.39 \\
        EfficientNetB3-R101-HVS & 13.84 & 16.14 & 42.50 & 124 & 76.41 & 56.13 & 87.07 & 74.53 & 62.72 & 36.86 & 71.54 & 49.09 \\
        \midrule
        R18-\modelname~(ours) & 11.44 & 28.62 & 11.44 & 28.62 & 81.71 & 59.73 & 90.67 & 80.15 & 74.05 & 46.48 & 77.24 & 57.90 \\
        R34-\modelname~(ours) & 21.55 & 57.71 & 21.55 & 57.71 & \textbf{83.05} & \textbf{64.16} & \textbf{92.33} & \textbf{83.20} & \textbf{76.88} & \textbf{52.65} & \textbf{80.57} & \textbf{63.26} \\
        \bottomrule
    \end{tabular}
    }
    \end{center}
    \caption{Comparison to approaches in asymmetric image retrieval on $\mathcal{R}$Oxf and $\mathcal{R}$Par datasets. Our model is distilled from R101-DOLG and R101-DELG. The best performance are marked in bold. GFLOPs of each method is estimated using an input image of size 1024$\times$768. Our method produces lightweight models for both query side and gallery side, which is more effcient than asymmetric image retrieval approaches using a heavy model for the gallery images. Our models are also superior than them in retrieval performance.}
    \label{tab:gldv2_asym}
\end{table*}
Recent asymmetric image retrieval methods \cite{wu2022contextual,budnik2021asymmetric,duggal2021compatibility} also aims for efficient image retrieval, which adopt separate models for query images and gallery images that query images are processed a lightweight model to maximize efficiency and gallery images are processed by a deep model to maximize performance. 
However, asymmetric image retrieval still suffers from heavy memory footprint and computation overhead for the offline gallery image feature extraction, especially in large-scale scenarios with billions of gallery images to process. Our distillation method produces single lightweight model for both the query side and gallery side which highlights the advantage of our approach against asymmetric image retrieval models. 
As in Table \ref{tab:gldv2_asym}, our model also achieves better performance than asymmetric image retrieval models.

\section{Training statistics}
\label{sec:stat}
\begin{figure}[tbp]
  \centering
  \begin{subfigure}{0.45\linewidth}
    \includegraphics[width=\linewidth]{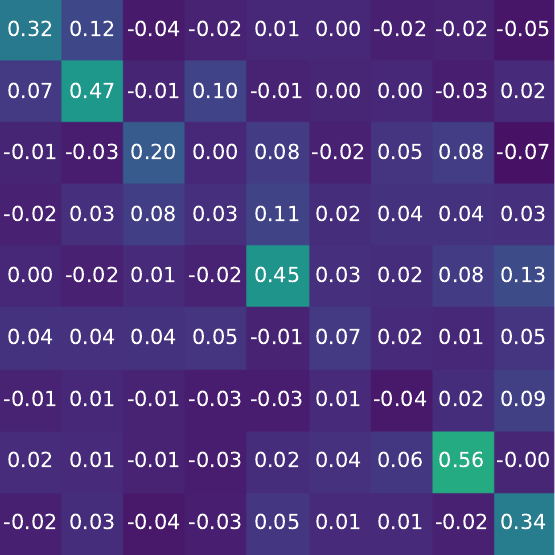}
    \caption{R101-GeM}
    \label{fig:wopca_sim_map_a}
  \end{subfigure}
  \begin{subfigure}{0.45\linewidth}
    \includegraphics[width=\linewidth]{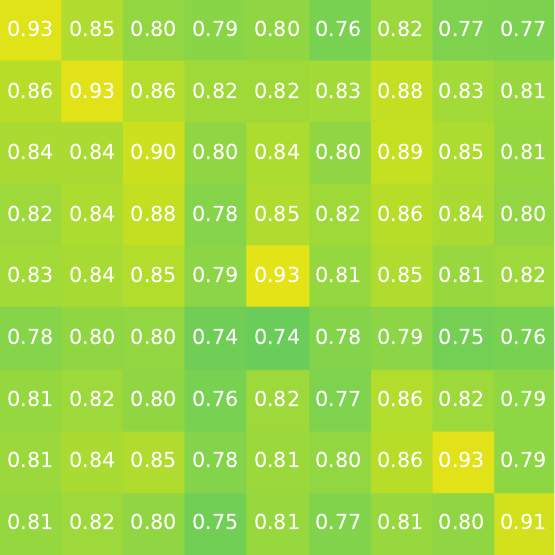}
    \caption{R101-AP-GeM}
    \label{fig:wopca_sim_map_b}
  \end{subfigure}

  \begin{subfigure}{0.45\linewidth}
    \includegraphics[width=\linewidth]{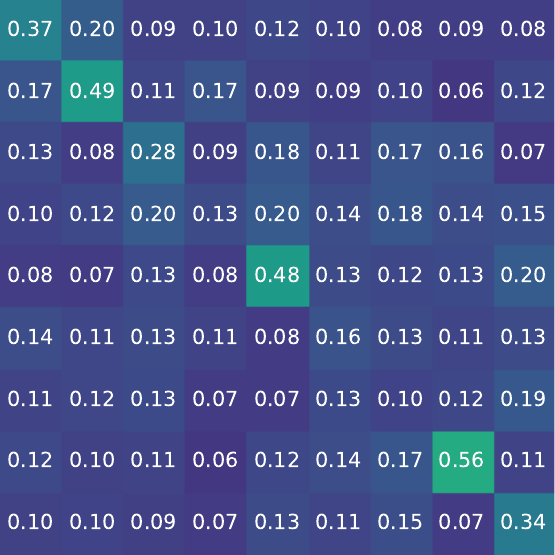}
    \caption{R101-SOLAR}
    \label{fig:wopca_sim_map_c}
  \end{subfigure}
  \begin{subfigure}{0.45\linewidth}
    \includegraphics[width=\linewidth]{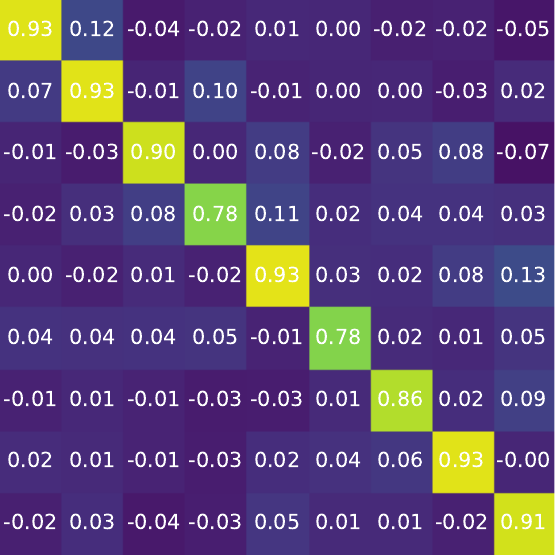}
    \caption{Fusion similarity matrix}
    \label{fig:wopca_sim_map_d}
  \end{subfigure}
  \caption{Example $9\times 9$ similarity matrices of R101-GeM, R101-AP-GeM, R101-SOLAR and their fusion similarity matrix using \textit{max-min} strategy \textbf{without PCA-whitening}. The similarity scores are depicted in each grid. Fusion on diagonal is dominated by R101-AP-GeM and the off-diagonal area is biased to R101-GeM and R101-SOLAR.
  }
  \label{fig:wopca_sim_map}
\end{figure}
\begin{figure}[tbp]
  \centering
  \begin{subfigure}{0.45\linewidth}
    \includegraphics[width=\linewidth]{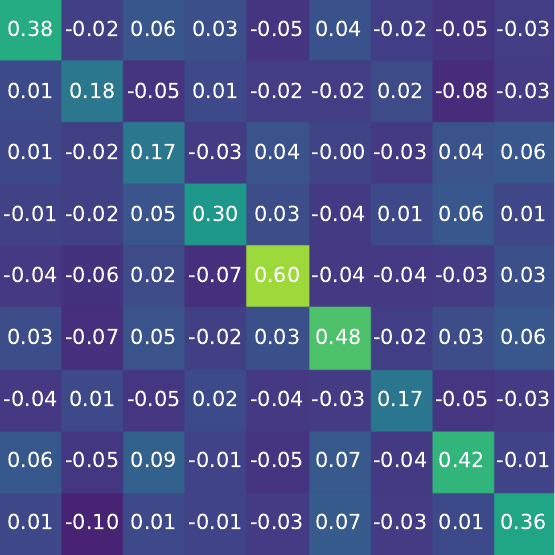}
    \caption{R101-GeM}
    \label{fig:sim_map_a}
  \end{subfigure}
  \begin{subfigure}{0.45\linewidth}
    \includegraphics[width=\linewidth]{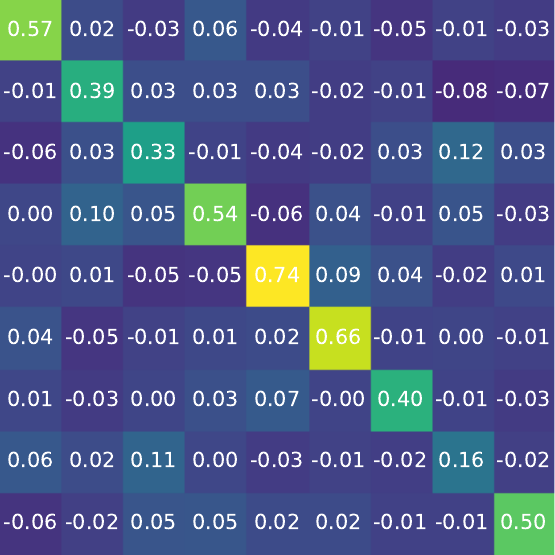}
    \caption{R101-AP-GeM}
    \label{fig:sim_map_b}
  \end{subfigure}

  \begin{subfigure}{0.45\linewidth}
    \includegraphics[width=\linewidth]{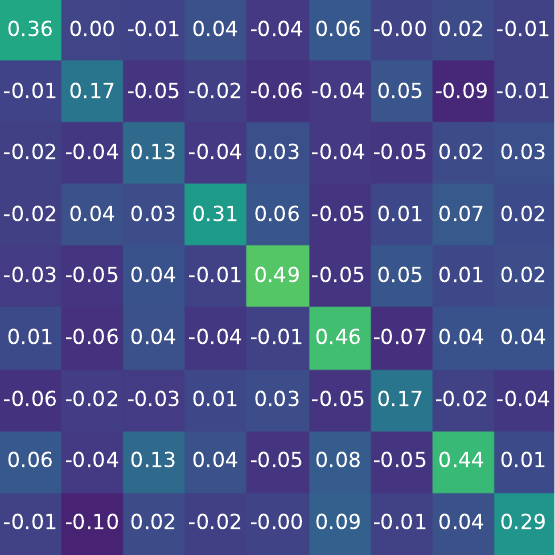}
    \caption{R101-SOLAR}
    \label{fig:sim_map_c}
  \end{subfigure}
  \begin{subfigure}{0.45\linewidth}
    \includegraphics[width=\linewidth]{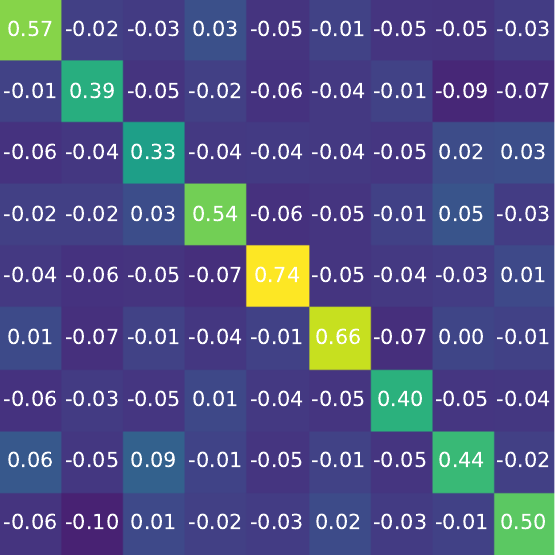}
    \caption{Fusion similarity matrix}
    \label{fig:sim_map_d}
  \end{subfigure}
  \caption{Example $9\times 9$ similarity matrices of R101-GeM, R101-AP-GeM, R101-SOLAR and their fusion similarity matrix using \textit{max-min} strategy \textbf{with PCA-whitening}. The similarity scores are depicted in each grid. The example demonstrates the complementarity between teacher models and the necessity of whitening teacher models.
  }
  \label{fig:sim_map}
\end{figure}
We further provide analysis on training two types of statistics to give a better understanding of our method, respective similarity scores and the mean reciprocal rank (MRR) during training.

\textbf{Similarity scores}. Example similarity matrices produced by teacher models R101-GeM, R101-AP-GeM and R101-SOLAR are depicted in Figure \ref{fig:wopca_sim_map} and Figure \ref{fig:sim_map}. Similarity scores in Figure \ref{fig:wopca_sim_map} are calculated using the original teacher models and Figure \ref{fig:sim_map} using whitend teachers. It can be seen that there is considerable discrepancy of similarity scores produced by original teacher models such that fusion of multiple teachers will be biased to certain teacher model, leading to sub-optimal distillation. After whitening, teacher models compete on the same level and their complementarity can be fully utilized.

\textbf{Mean reciprocal rank (MRR)}. We use MRR as a measure of the quality of a similarity matrix $\mathcal{M}\in \mathbb{R}^{N\times N}$, where N is the batch size and the similarity on diagonal of $\mathcal{M}$ are the similarity of positive pairs. Then MRR of $\mathcal{M}$ can be calculated as:
\begin{equation}
    MRR(\mathcal{M}) = \frac{1}{N}\sum_{i=1}^N \frac{1}{rank_i},
\end{equation}
where $rank_i$ refers to the rank of positive pair $(x_i, y_i)$ in the descending ordered list of $\mathcal{M}[i, :]$. A greater MRR indicates that the similarity predicted for positive pairs is more prominent among all the similarities.

\begin{figure}[t]
  \centering
    \includegraphics[width=0.9\linewidth]{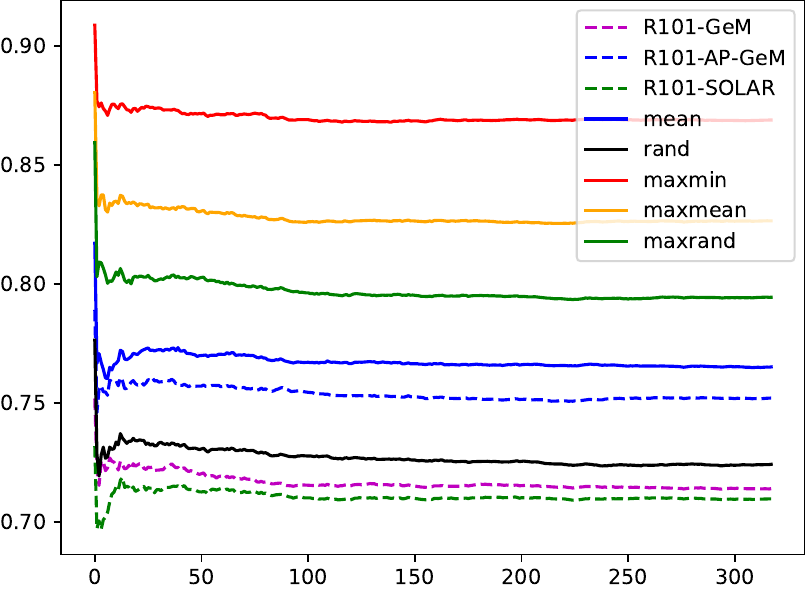}
  \caption{Cumulative average MRR of one epoch on GLDv2-clean. MRRs of different fusion strategies are calculated in triple-teacher distillation. Overall, MRR metric in training coincides with the evaluation performance. MRRs produced by multi-teacher fusion are better than single teacher models, demonstrating the effectiveness of our multi-teacher aggregation mechanism.}
  \label{fig:mrr}
\end{figure}

Figure \ref{fig:mrr} presents the cumulative average MRR of different models during distillation for one epoch of total 318 mini-batches. We draw the cumulative average MRR to eliminate jitter between mini-batches. As shown, the fusion similarity matrices using \textit{max-min} strategy has the largest MRR, consistent to its best performance among all the strategies. Compared to MRR of single teacher model, fusion similarity matrices produce higher MRRs which demonstrates the effectiveness of our multi-teacher fusion approach by aggregating similarity matrices of teacher models.

\section{Implementation details}
\label{sec:imple_details}

\subsection{Landmark image retrieval}
\subsubsection{Teacher models}
We collect five pre-trained landmark image retrieval networks for multi-teacher distillation, respectively R101-GeM \cite{radenovic2018fine}, R101-AP-GeM \cite{revaud2019learning}, R101-SOLAR\cite{ng2020solar}, R101-DELG \cite{cao2020unifying,yang2021dolg} and R101-DOLG \cite{yang2021dolg}. To adapt them for distillation, we made some modifications as follows.

\textbf{R101-GeM} \cite{radenovic2018fine} in the original publication ends up with no fully connected~(fc) layer but learns a whitening transformation in the post-processing process. We use their later released version (of best performance) in their github repository trained on GLDv1 which appends a fully connected (fc) layer to the ResNet backbone and learns it end-to-end. We fix the parameter $p$ in GeM pooling layer in distillation. Note that there is an additional $l_2$-normalization layer added before the last fc layer.

\textbf{R101-AP-GeM} \cite{revaud2019learning} has similar network structure to R101-GeM except that it is trained via AP loss. The parameter $p$ in GeM pooling is learned along with other parameters, we fix it to 3 instead. We use their released pre-trained weights trained on GLDv1.

\textbf{R101-SOLAR} \cite{ng2020solar} utilizes second-order (self) attention to re-weight spatial features. We fix $p$ in GeM pooling to 3. R101-SOLAR is trained on GLDv1.

\textbf{R101-DELG} \cite{cao2020unifying,yang2021dolg} consists of global features and local features where local features are used to perform re-ranking. We only utilize its global features in our experiments. Parameter $p$ in GeM pooling is also fixed. We use the version re-implemented by \cite{yang2021dolg} trained on GLDv2-clean.

\textbf{R101-DOLG} \cite{yang2021dolg} fuses global features and local features into unified descriptors. We use its global outputs. Parameter $p$ in GeM pooling is fixed. R101-DOLG is trained on GLDv2-clean.

GLDv2-clean is used for training PCA-Whitening. The image features extracted by teacher models are first $l_2$-normalized. PCA-whitening is learned on the $l_2$-normalized features. Another $l_2$-normalization is finally performed after whitening. Input preprocessing is the same as the student model in distillation, see Section \ref{sec:stu_implementation}.

\subsubsection{Student model}
\label{sec:stu_implementation}
The student model is initialized with pre-trained ImageNet weights and fine-tuned in distillation. The original linear layer for classification at the end of the model is discarded and we append a new randomly initialized linear layer to adjust its embedding dimension. The average pooling layer at the second last layer is replaced by the Generalized Mean (GeM) pooling \cite{radenovic2018fine} with a fixed parameter $p$ as 3. 

\textbf{Training details}. During training, Adam optimizer is adopted with an initial learning rate of $10^{-3}$ and a weight decay of $10^{-6}$. The learning rate is decayed with a cosine annealing scheduler. Temperature $\tau$ in the loss function is set to 0.05 for both teachers and student. At each iteration, we randomly sample $N$ landmark instances and two images from each instance. We randomly crop and resize each image to $512\times 512$ and perform random horizontal flip. The student model is trained with batch size of 256 on 8 V100 GPUs with 16G memory per card. 

\textbf{Evaluation details}. At inference time, we refer to the common practice in the field \cite{radenovic2018fine,revaud2019learning,cao2020unifying} to perform retrieval evaluation. We extract multi-scale features of 3 scales $\left\{ 1/2,1/\sqrt{2},1 \right\}$. Features of each scale is $l_2$-normalized first, then averaged and $l_2$-normalized again. For a clear verification of our distillation method, we only evaluate the retrieval performance based on global features, without any query expansion \cite{radenovic2018fine,revaud2019learning} or re-ranking procedure \cite{cao2020unifying}. Note that we evaluate the teacher models under the same setting as ours (refer to supplementary materials) for a fair comparison, considering the various evaluation settings of them (\textit{e.g.}, query crop, scales of multi-scale feature extraction). Therefore the performance of teacher models in Table 1 might be different from their original reports.

\subsection{Near-duplicate video retrieval}
\subsubsection{Datasets}
We utilize Short Video Dataset (SVD)~\cite{jiang2019svd}, considering it is the latest and the largest benchmark dataset for near-duplicate video retrieval.
It contains 562,013 short videos, and all videos are collected from the short video sharing platform Douyin (a Chinese version of TikTok).
Among them, there are 1,206 query videos and 34,020 labeled videos which are either positives (near-duplicate) or hard negatives of the query videos.
The rest of videos are unlabeled, which are probable negative videos to the query videos.
We use the official train-test split in \cite{jiang2019svd}, with 1000 query videos for training and the rest for testing. 
For each test query video, its corresponding labeled set and the whole unlabeled set are utilized as candidates to rank.

\subsubsection{Teacher models} 
Recent self-supervised learning models are trained to be invariant to image transformations, which is more suitable for near-duplicate video retrieval than the widely used classification pre-trained models. Therefore, we collect two self-supervised pre-trained models as teachers, respectively R50-MoCov3 \cite{chen2021empirical}, R50-BarlowTwins \cite{zbontar2021barlow}. Projection head or prediction head in these models are discarded and only the output of R50 backbone is used for distillation.

\subsubsection{Student model}
The student network in near-duplicate video retrieval task is the same as that in landmark image retrieval, except for that we do not replace the original average pooling layer.

\subsubsection{Training details}
At training time, we extract video frames at a sampling rate of 1 fps. SVD's training labeled set is used for distillation. At each iteration, $N$ videos are randomly sampled and then a frame is randomly sampled from each video. We follow the data augmentation approaches used in recent self-supervised learning models \cite{grill2020bootstrap,chen2021empirical,zbontar2021barlow,caron2021emerging} to generate positive pairs. 

\subsubsection{Evaluation details}
At inference time, shorter side of each video frame are resized to 256, preserving its aspect ratio, then a $224\times 224$ patch is cropped in the center. No additional augmentations are applied. The extracted frame features are $l_2$-normalized and used to calculate a frame-to-frame similarity matrix between two videos. We adopt Chamfer Similarity \cite{kordopatis2019visil} to aggregate the similarity matrix into a single similarity value, which is finally used to rank the candidate videos.

\bibliography{supp/ref}